\documentclass[journal]{IEEEtran}

\usepackage{ifpdf}
\usepackage{cite}
\usepackage{amsmath}
\usepackage[mathcal]{eucal}

\usepackage{gensymb}
\usepackage{epsfig} 
\usepackage{algorithm}   
\usepackage{algpseudocode} 
\usepackage{adjustbox}
\usepackage{array}
\usepackage{mathtools}
\usepackage{colortbl}
\usepackage{commath}
\usepackage{mdwmath}
\usepackage{eqparbox}
\usepackage{bbding}
\usepackage{booktabs}
\usepackage{url}
\usepackage{multirow,array}
\usepackage{graphicx}
\usepackage{bm}
\usepackage{textcomp} 
\usepackage{amssymb} 
\usepackage{threeparttable}

\RequirePackage{ifpdf}
\ifpdf \PassOptionsToPackage{colorlinks=true}{hyperref} \fi
\RequirePackage{hyperref}
\hypersetup{linkcolor=black,citecolor=green}

\ifCLASSOPTIONcompsoc
\usepackage[caption=false, font=normalsize, labelfont=sf, textfont=sf]{subfig}
\else
\usepackage[caption=false, font=footnotesize]{subfig}
\fi

\graphicspath{{./results/}}
\DeclareGraphicsExtensions{.pdf,.jpeg,.png,.jpg}
   
\IEEEoverridecommandlockouts                             
\usepackage{graphics} 
\usepackage{epsfig} 
\usepackage{mathptmx} 
\usepackage{times} 
\usepackage{mdwmath}

\DeclareMathOperator*{\argmin}{arg\,min}

\begin{document}
\renewcommand{\arraystretch}{1.3}
\title{MIL-LC: A Robust Magnetometer-Inertial-LiDAR Fusion Multimodal Localization Framework}

\author{

Qiyang Lyu$^{*}$, Zhenyu Wu$^{*}$, Wei Wang, Hongming Shen, and Danwei Wang,~\IEEEmembership{Life Fellow,~IEEE}

\vspace{-1em}

\thanks{ 

This research is supported by the National Research Foundation, Singapore under its Medium Sized Centre for Advanced Robotics Technology Innovation (CARTIN). (\textit{Corresponding author: Danwei Wang})

The authors are with the School of Electrical and Electronic Engineering, Nanyang Technological University, Singapore 639798 (e-mail: \{qiyang001, zhenyu002, wei013\}@e.ntu.edu.sg; \{hongming.shen, edwwang\}@ntu.edu.sg).

$^{*}$Co-first authorship
}}

\markboth{MANUSCRIPT PREPARED FOR IEEE Transactions on Industrial Electronics}%
{}

\maketitle

\begin{abstract}
Localization in challenging environments, such as GNSS-denied, geometrically repetitive, or textureless scenes commonly found in offices, hotels, and underground parking facilities, remains an open problem for reliable autonomous mobile robot (AMR) deployment. Single-modality localization methods are inherently limited by the constraints of individual sensors. Although multimodal fusion frameworks have shown improved robustness, most existing approaches still rely heavily on geometric or texture features, or on infrastructure-based beacons, which increase installation and maintenance costs while reducing deployment flexibility. Recently, ambient magnetic field (AMF)-based localization has attracted growing attention because it does not depend on geometric or texture features, nor does it require additional infrastructure, making it a promising complementary modality for AMR localization. However, existing studies have only explored such fusion in pedestrian scenarios using smartphone-mounted sensor suites, and practical solutions for AMR systems remain largely unexplored. To address this gap, this article proposes a magnetometer-inertial-LiDAR fused multimodal localization framework with a custom-designed sensor suite, termed MIL-LC, which provides reliable localization even when LiDAR suffers from geometric degeneration or when the magnetic map changes during long-term deployment. Extensive experiments in both simulation and real-world environments demonstrate that the proposed MIL-LC framework achieves robust and accurate localization performance.
\end{abstract}

\section{Introduction}
The widespread deployment of robotics across diverse application scenarios \cite{wu2026mc, li2025simsl} has increased the demand for robots to perceive, understand, and interact with their environments reliably before executing downstream tasks. These scenarios range from relatively structured environments, where abundant geometric and texture information supports localization \cite{VINS-Mono, Fast-LIO}, to highly challenging settings, such as geometrically degenerate or repetitive environments for LiDAR and cameras \cite{wu2024mglt}, GNSS-denied environments \cite{he2020integrated}, and obstacle-cluttered environments that degrade UWB-based localization. Such conditions are common in real-world applications, yet robots are still expected to operate robustly within them. Under these circumstances, localization methods that rely on a single sensor often have limited generalizability, making multi-sensor fusion a more practical solution. Existing studies have explored various multimodal fusion strategies \cite{xu2024lcdl, debeunne2020review, nguyen2021range, wu2023rf, he2020integrated}, improving robustness by leveraging complementary sensing modalities. However, most existing fusion frameworks still rely heavily either on environmental geometric or texture information, or on external infrastructure such as beacons, which restricts deployment flexibility and increases maintenance cost.
\begin{figure}[t]
	\centering
	\includegraphics[width=0.9\linewidth]{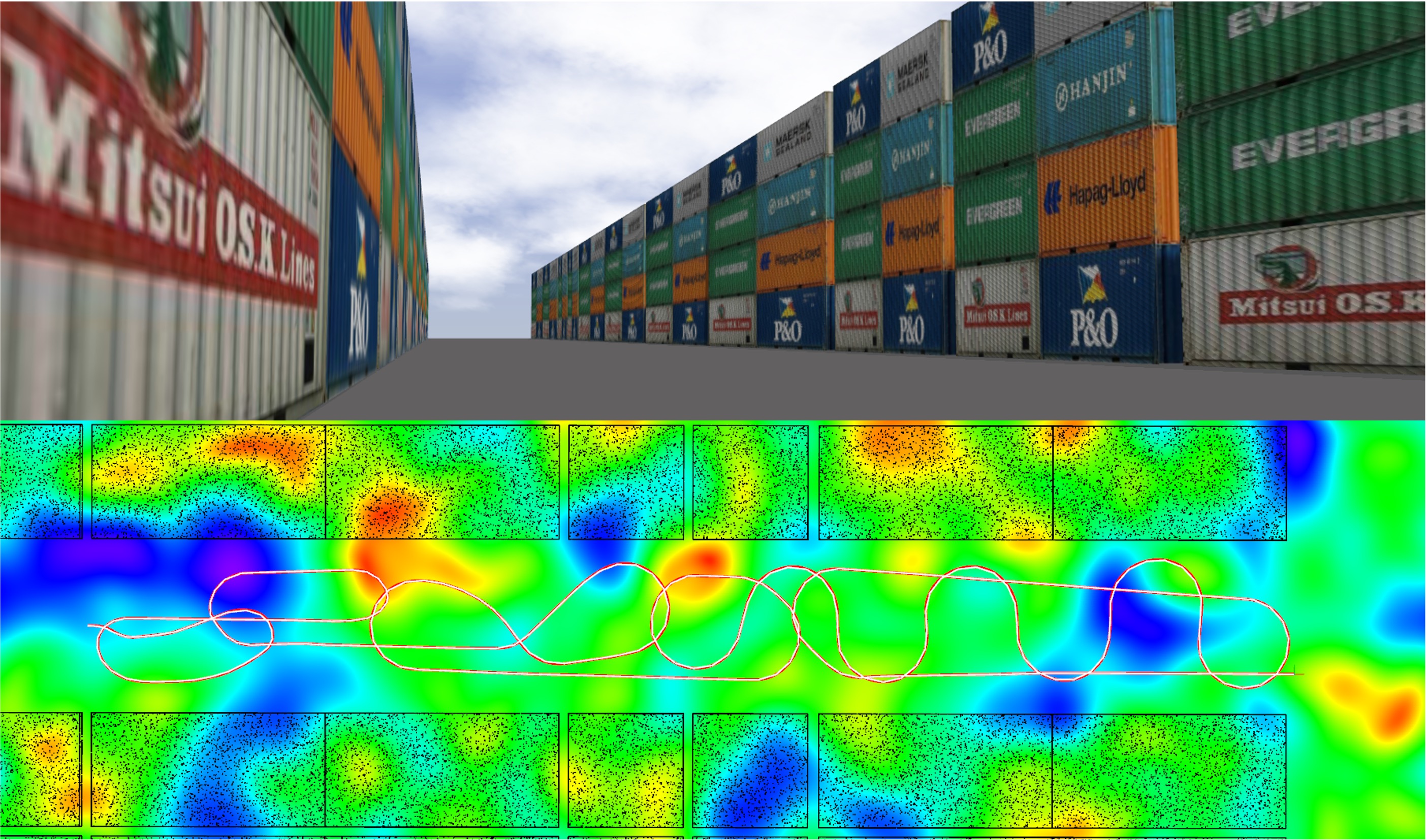}
	\caption{The high-fidelity simulated seaport environments and the localization results using our proposed MIL-LC method (white) compared with the ground truth path (red).}
	\label{fig:simulated_seaport}
	\vspace{-1em}
\end{figure}
Inspired by the ability of birds to navigate using the Earth’s magnetic field \cite{mouritsen2018long}, the ambient magnetic field (AMF) has emerged as a promising modality for robust and flexible fusion-based localization \cite{wu2026mc, wu2024mglt}. Distorted by nearby ferromagnetic objects, the AMF exhibits location-dependent patterns that can serve as fingerprints \cite{li2012feasible}. Such magnetic signatures are commonly found in enclosed or semi-enclosed environments, including office and hospital corridors, supermarkets, and car parks because of the widespread presence of built-in ferromagnetic objects (e.g., steel bars, power cables, pipes). Previous studies have shown that magnetic-based localization can provide localization cues similar to GNSS \cite{idf_mfl, subbu2013locateme}. However, single-modal magnetic-based localization usually suffers from limited accuracy. Existing multimodal methods using magnetic information, such as radio-magnetic and visual-magnetic fusion \cite{jung2015indoor, magicol2015, WangMag2019, vmag2017, yang2025multi}, are still mainly restricted to pedestrian and smartphone applications. In contrast, magnetic fusion for AMRs localization, particularly with LiDAR, remains largely unexplored, and its long-term robustness has yet to be sufficiently validated.

To address these limitations, this article proposes a loosely coupled and easy-to-implement magnetometer-inertial-LiDAR multimodal localization framework. The proposed framework can adaptively balance localization accuracy and robustness according to environmental conditions, while also maintaining long-term stability. The main contributions are summarized as follows:

\begin{itemize}
	\item A novel multimodal localization framework named MIL-LC that fuses magnetic and LiDAR information while incorporating inertial measurements for AMRs' robust localization in challenging degenerate environments.
	\item A newly-designed novel standalone sensor suite with multiple complementary modalities and onboard computing unit.
	\item Adaptive multimodal balancing strategies, including degeneracy detection, outlier rejection, and adaptive weighting, to improve the system's robustness and adaptability under varying environmental conditions.
	\item Extensive experiments in both simulated and real-world environments demonstrate strong robustness and accuracy across various scenarios, including long-term resilience to magnetic disturbance.
\end{itemize}

The remainder of this article is organized as follows. Section \ref{sec:lr} reviews related work. Section \ref{sec:model} and \ref{sec:method} build up the system model and presents the proposed methodology, including undistortion and state update. Section \ref{sec:chapter6/exp} evaluates the framework through extensive simulated and real-world experiments. Finally, Section \ref{sec:chapter6/conclusion} concludes this article.


\section{Literature Review} \label{sec:lr}
Multimodal localization is motivated by the complementary failure modes of individual sensors. Cameras are sensitive to illumination change, occlusion, and weak texture; LiDAR can degenerate in geometrically repetitive environments; IMUs drift over time; GNSS is vulnerable to blockage and multipath; UWB suffers under NLOS conditions and poor anchor geometry; and RFID often depends on prior tag deployment or external references. Fusing multiple modalities is therefore a natural strategy for improving localization robustness.

Conventional multimodal fusion systems often combine vision with geometric sensing, particularly LiDAR, to compensate for the limitations of pure vision-based localization. In such systems, LiDAR improves robustness under poor illumination and weak texture, while cameras enhance target association \cite{xu2024lcdl, debeunne2020review}. Other studies further integrate infrastructure-based positioning sensors. Camera-UWB fusion has been explored through constrained filtering and tightly coupled optimization to reduce long-term drift \cite{hao2025information,nguyen2021range}. LiDAR-GNSS fusion commonly switches between GNSS-assisted and LiDAR-dominant modes, applies confidence-aware strategies to smooth transitions under varying GNSS quality, or incorporates adaptive filtering and graph optimization to mitigate drift in urban environments \cite{he2020integrated,gao2023gnss,shen2024lidar}. LiDAR-IMU-UWB fusion has also shown strong performance in GNSS-denied or geometrically degenerate settings such as tunnels, where LiDAR is weakly observable along the tunnel axis and UWB provides complementary absolute ranging \cite{zhou2020lidar, kuang2024tightly}. Beyond these mainstream combinations, RFID-based fusion has been investigated for warehouse localization, where RFID phase and odometry constraints are jointly optimized in a graph framework \cite{wu2023rf}. Overall, multimodal fusion generally improves localization accuracy and robustness, but at the cost of higher calibration effort, greater system complexity, increased computation, and stronger dependence on scenario-specific assumptions such as anchor deployment and maintenance \cite{fan2025lidar}.

On the other hand, early studies on magnetic-based localization as a standalone modality demonstrated its feasibility, but also revealed inherent limitations, including limited spatial discriminability of magnetic signals and low signal-to-noise ratio, which generally lead to lower localization accuracy \cite{subbu2013locateme,WangMag2019,magicol2015}. To address these limitations, increasing attention has been given to fusing magnetic sensing with other modalities. Jung et al. \cite{jung2015indoor} combined magnetic and radio measurements within a Rao-Blackwellized particle filter SLAM framework, improving global relocalization performance. VMag \cite{vmag2017} further integrated geomagnetic and visual information through a context-aware particle filter, where magnetic measurements provided locally stable signatures while visual features contributed richer spatial constraints. MaLoc \cite{WangMag2019} and Magicol \cite{magicol2015} also demonstrated that combining magnetic observations with auxiliary cues such as pedestrian motion information or WiFi can improve localization robustness. More recently, magnetic measurements are fused with inertial motion estimates on a magnetic map, enabling large-scale infrastructure-free localization with stronger robustness \cite{liuMagloc2023, deshpande2024magnetic}. Yang et al. \cite{yang2025multi} built up a multi-sensor fusion system with IMU, magnetometer, Bluetooth and so on, but the magnetic information is only partially utilized for heading estimation.

In summary, although these studies indicate that magnetic sensing is particularly valuable as a complementary modality, most of them still depend on sequential matching or particle-filter-style inference, which can be computationally expensive and often requires sufficient temporal history or a constrained search space to avoid mismatching, and many fusion systems remain closer to pedestrian or smartphone localization than to tightly coupled robot localization. This paper therefore focuses on addressing this limitation by investigating a fusion framework for robust localization for AMRs.

\section{System Modeling}\label{sec:model}
\begin{figure}[t]
	\centering
	\includegraphics[width=0.99\linewidth]{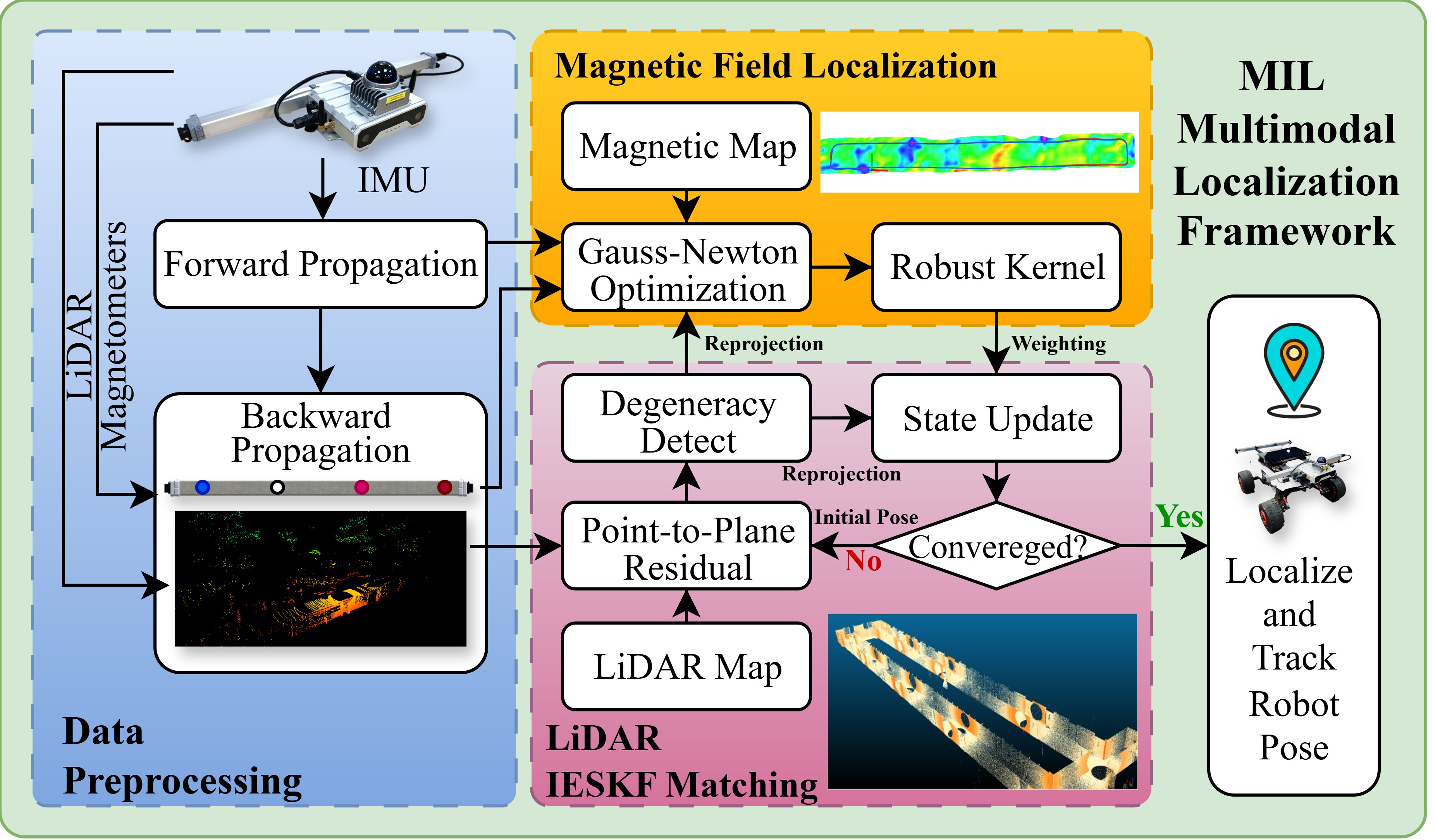}
	\caption{Flowchart of the proposed framework. The overall framework can be divided into three parts: (a) Data synchronization and motion undistortion, (b) Magnetic-based localization and robust kernel filtering, and (c) LiDAR-based degeneration detection and IESKF update.}
	\label{fig:ch6_diagram}
	\vspace{-1em}
\end{figure}

The overall framework of the multimodal localization system is illustrated in Fig.~\ref{fig:ch6_diagram}. Measurements from different sensors are first synchronized and corrected for motion distortion through IMU-based forward and backward propagation, as indicated in the blue block. The preprocessed data are then fed into their respective localization modules for pose estimation, as shown in the pink and orange blocks. Notably, the pose estimate obtained from the magnetometer is utilized as the initial guess for LiDAR scan matching. After the LiDAR-based iterated error-state Kalman filter (IESKF) converges, the refined pose estimate is output as system odometry.
\vspace{-1em}

\subsection{Kinematic Model}
In the kinematic model, the IMU frame is adopted as the body frame for state estimation, and a state vector and kinematic model are employed. In this work, the global frame is aligned with the coordinate frame of the pre-built global map. Accordingly, the IMU-based kinematic model is defined as follows:
\vspace{-0.5em}
\begin{equation}
	\small{
	\begin{aligned}
		^{\mathcal{M}}\dot{\mathbf{R}}_{I} &=  {^{\mathcal{M}}\mathbf{R}_{I}}\lfloor\boldsymbol{\omega}_{m} - \mathbf{b}_{\boldsymbol{\omega}} - \mathbf{n}_{\boldsymbol{\omega}}\rfloor_{\wedge}, ^{\mathcal{M}}\dot{\mathbf{v}}_{I} = {^{\mathcal{M}}\mathbf{R}_{I}}(\mathbf{a}_{m} - \mathbf{b}_{\mathbf{a}} - \mathbf{n}_{\mathbf{a}}) + {^{\mathcal{M}}\mathbf{g}}\\
		^{\mathcal{M}}\dot{\mathbf{p}}_{I} &= {^{\mathcal{M}}\mathbf{v}_{I}} ,\, 
		\dot{\mathbf{b}}_{\boldsymbol{\omega}} = \mathbf{n}_{\mathbf{b}\boldsymbol{\omega}},\,
		\dot{\mathbf{b}}_{\mathbf{a}} = \mathbf{n}_\mathbf{ba}, \,
		^{\mathcal{M}}\dot{\mathbf{g}} = \mathbf{0}
	\end{aligned}
}
\vspace{-0.5em}
\end{equation}
where ${^{\mathcal{M}}\mathbf{R}_{I}}$, ${^{\mathcal{M}}\mathbf{p}_{I}}$ and  ${^{\mathcal{M}}\mathbf{v}_{I}}$ are the orientation position, and linear velocity of IMU in the global map frame, and ${^{\mathcal{M}}\mathbf{g}}$ is the gravity vector, $\mathbf{n}_{\boldsymbol{\omega}}$ and $\mathbf{n}_{\mathbf{a}}$ are the measurement noise of accelerometer and gyroscope, $\mathbf{b}_{\mathbf{a}}$ and $\mathbf{b}_{\boldsymbol{\omega}}$ are the IMU measurement bias with the random walk as $\mathbf{n}_\mathbf{ba}$ and $\mathbf{n}_{\mathbf{b}\boldsymbol{\omega}}$. Finally, $\lfloor\cdot\rfloor$ denotes the skew-symmetric cross product. 

According to the IMU kinematic model, the system state vector $\mathbf{x}$, input vector $\mathbf{u}$, and noise vector $\mathbf{w}$ can be defined as:
\vspace{-0.5em}
\begin{equation}
	\small{
	\begin{aligned}
		\mathbf{x} &= \left[{^{\mathcal{M}}\mathbf{R}_{I}^{\top}}\ {^{\mathcal{M}}\mathbf{p}_{I}^{\top}}\ {^{\mathcal{M}}\mathbf{v}_{I}^{\top}}\ \mathbf{b}_{\boldsymbol{\omega}}^{\top}\ \mathbf{b}_{\mathbf{a}}^{\top}\ {^{\mathcal{M}}\mathbf{g}^{\top}}\right]^{\top} \\
		\mathbf{u} &= \left[\boldsymbol{\omega}_{m}^{\top}\ \mathbf{a}_{m}^{\top}\right]^{\top}, \,
		\mathbf{w} = \left[\mathbf{n}_{\boldsymbol{\omega}}^{\top}\ \mathbf{n}_{\mathbf{a}}^{\top}\ \mathbf{n}_{\mathbf{b}\boldsymbol{\omega}}^{\top}\ \mathbf{n}_\mathbf{ba}^{\top}\right]^{\top}
	\end{aligned}
}
\vspace{-0.5em}
\end{equation}

\subsection{State Transition and Forward Propagation}
\begin{figure}[!t]
	\centering
	\includegraphics[width=0.93\linewidth]{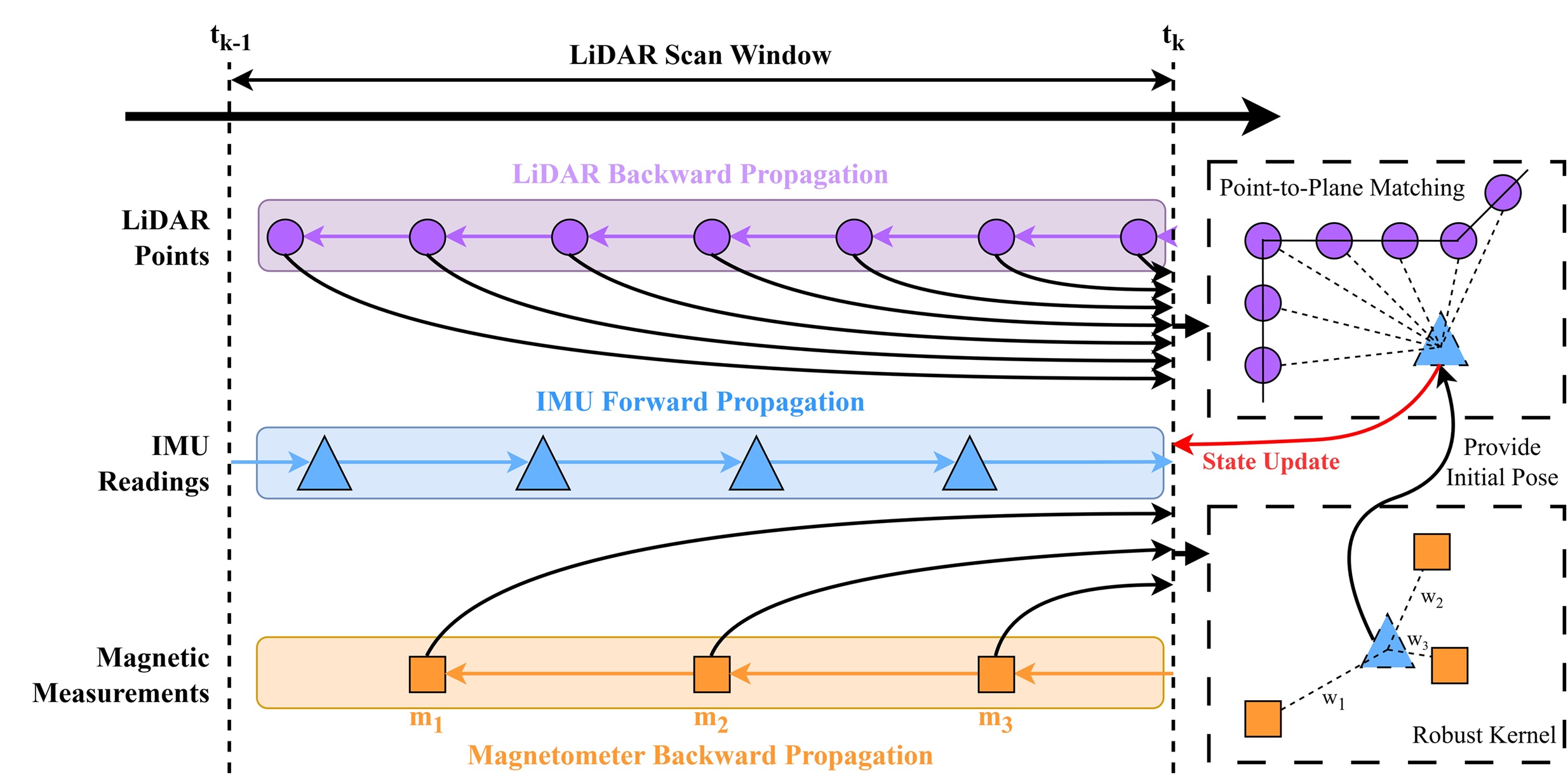}
	\caption{Data synchronization and motion undistortion of magnetic and LiDAR data using forward and backward propagation during the state prediction.}
	\label{fig:ch6_propagation}
	\vspace{-1em}
\end{figure}

Before incorporating the sensor measurements, the system first propagates the state from $t_{k-1}$ to $t_k$ using IMU readings. Based on the discretization of the continuous-time kinematic model with an IMU sampling interval of $\Delta t$, the state propagation can be expressed as follows:
\vspace{-0.5em}
\begin{equation}
	\label{eq:forward_propagation}
	\small{
	\hat{\mathbf{x}}_{t_i} = \hat{\mathbf{x}}_{t_{i-1}}\boxplus
	\begin{bmatrix}
		(\boldsymbol{\omega}_{m}-\mathbf{b}_{\boldsymbol{\omega}}-\mathbf{n}_{\boldsymbol{\omega}})\Delta t \\
		{^{\mathcal{M}}\mathbf{v}_{I}}\Delta t + \frac{1}{2}({^{\mathcal{M}}\mathbf{R}_{I}}(\mathbf{a}_{m} - \mathbf{b}_{\mathbf{a}} - \mathbf{n}_{\mathbf{a}}) + {^{\mathcal{M}}\mathbf{g}})\Delta t^2 \\
		((\mathbf{a}_{m} - \mathbf{b}_{\mathbf{a}} - \mathbf{n}_{\mathbf{a}}) + {^{\mathcal{M}}\mathbf{g}})\Delta t \\
		\mathbf{n}_{\mathbf{b}\boldsymbol{\omega}}\Delta t \\
		\mathbf{n}_\mathbf{ba}\Delta t\\
		\mathbf{0}_{3\times1}
	\end{bmatrix}
}
\vspace{-0.5em}
\end{equation}
Between two consecutive state updates, multiple IMU measurements may be received. Starting from the last updated state, $\hat{\mathbf{x}}_{t_{0}} = \bar{\mathbf{x}}_{t_{k-1}}$, the state is propagated using these IMU readings until reaching the end time of the scan window. This forward propagation is illustrated in Fig.~\ref{fig:ch6_propagation} as the blue triangles between two dashed lines. During this stage, the state covariance accumulates due to the absence of observations: $\hat{\mathbf{P}}_{t_{i}} = \mathbf{F}_{\tilde{\mathbf{x}}_{t_{i-1}}}\hat{\mathbf{P}}_{t_{i-1}}\mathbf{F}_{\tilde{\mathbf{x}}_{t_{i-1}}}^{\top}+\mathbf{F}_{\mathbf{w}_{t_{i-1}}}\mathbf{Q}_{t_{i-1}}\mathbf{F}_{\mathbf{w}_{t_{i-1}}}^{\top}$
where $\mathbf{Q}_{t_{i-1}}$ is the covariance of noise. If the IMU is properly manufactured, the random walks can be neglected by setting $\mathbf{w} = \mathbf{0}$, and $\mathbf{F}_{\tilde{\mathbf{x}}_{t_{i-1}}}$ and $\mathbf{F}_{\mathbf{w}_{t_{i-1}}}$ can be calculated as 
\vspace{-0.5em}
\begin{equation}
	\small{
	\begin{aligned}
		\mathbf{F}_{\tilde{\mathbf{x}}_{t_{i-1}}} &= \left.\frac{\partial\delta\hat{\mathbf{x}}_{t_i}}{\partial\delta\mathbf{x}_{t_{i-1}}}\right|_{\tilde{\mathbf{x}}_{t_{i-1}}=\mathbf{0}, \mathbf{w}_{t_{i-1}}=\mathbf{0}}, \ \
		\mathbf{F}_{\mathbf{w}_{t_{i-1}}} &= \left.\frac{\partial\delta\hat{\mathbf{x}}_{t_i}}{\partial\mathbf{w}_{t_{i-1}}}\right|_{\tilde{\mathbf{x}}_{t_{i-1}}=\mathbf{0}, \mathbf{w}_{t_{i-1}}=\mathbf{0}}
	\end{aligned}
}
\vspace{-0.5em}
\end{equation}

\section{Methodology of Multimodal Localization}\label{sec:method}
\subsection{LiDAR Undistortion and Localization} 
We adopt the widely accepted framework FAST-LIO2 \cite{fast_lio2} as the backbone for LiDAR localization with some modifications. Since LiDAR systems may acquire points sequentially, the sampling time of each point will differ from the scan end time $t_k$. As a result, platform motion during the scan can introduce distortion into the full point cloud. To improve scan quality, this distortion is compensated through backward propagation of the predicted state. As illustrated in Fig.~\ref{fig:ch6_propagation}, the backward propagation is initialized from the end of the scan window.
\vspace{-0.5em}
\begin{equation}
	\label{eq:backward_propagation}
	\small{
	\begin{aligned}
		{^{I_{t_k}}\check{\mathbf{v}}_{I_{t_{i-1}}}} &= {^{I_{t_k}}\check{\mathbf{v}}_{I_{t_{i}}}} - {^{I_{t_k}}\check{\mathbf{R}}_{I_{t_{i}}}}(\mathbf{a}_{m_{t_{i-1}}} - \hat{\mathbf{b}}_{\mathbf{a}_{t_k}})\Delta{t} - {^{I_{t_k}}\hat{\mathbf{g}}_{t_k}}\Delta{t} \\
		{^{I_{t_k}}\check{\mathbf{p}}_{I_{t_{i-1}}}} &= {^{I_{t_k}}\check{\mathbf{p}}_{I_{t_{i}}}} - {^{I_{t_k}}\check{\mathbf{v}}_{I_{t_{i}}}}\Delta{t}, \ \ \
		{^{I_{t_k}}\check{\mathbf{R}}_{I_{t_{i-1}}}} = {^{I_{t_k}}\check{\mathbf{R}}_{I_{t_{i}}}}\exp((\hat{\mathbf{b}}_{\boldsymbol{\omega}_{t_k}}-\boldsymbol{\omega}_{m_{t_{i-1}}})\Delta{t})
	\end{aligned}
}
\vspace{-0.5em}
\end{equation}
where ${^{I_{t_k}}\check{\mathbf{T}}_{I_{t_i}}} = ({^{I_{t_k}}\check{\mathbf{R}}_{I_{t_{i}}}}, {^{I_{t_k}}\check{\mathbf{p}}_{I_{t_{i}}}})$ is the desired relative transformation between time $t_i$ and $t_k$. Then the LiDAR point can be synchronized from $t_{i}$ to $t_k$ with $	{^{L}\mathbf{p}_{t_k}} = {^{I}\mathbf{T}_{L}^{-1}}{^{I_{t_k}}\check{\mathbf{T}}_{I_{t_i}}}{^{I}\mathbf{T}_{L}}{^{L}\mathbf{p}_{t_i}}$
where ${^{I}\mathbf{T}_{L}}$ denotes the extrinsic transformation from the LiDAR frame to the IMU frame, as the LiDAR scan points are expressed in the LiDAR body frame rather than the IMU body frame. After all points are undistorted to a common coordinate frame at the scan end time, a complete undistorted LiDAR scan is obtained. Different from the scan-to-local matching strategy in \cite{fast_lio2}, a scan-to-global approach is adopted in this work since a global map is pre-built. To reduce computational cost, a local sliding-window map is maintained, where an initial local map cube is constructed and centered at the current body pose, and is shifted when the body pose approaches the boundary to ensure that the body pose remains near the center. Subsequently, the LiDAR scan points are aligned with the global map using a point-to-plane fitting strategy, and more detailed formulations of the residual computation and the iterated state update between the LiDAR scan and the local map can be found in \cite{fast_lio2}.
\vspace{-1em}
\subsection{Magnetometer Synchronization and Localization}
We propose a localization method for magnetic-based localization. Due to the high-frequency output of the magnetometer ($\approx 200Hz$), multiple measurements may be received within a single LiDAR scan window. Therefore, a similar backward propagation procedure is applied to the magnetometer observations to synchronize them to a common time frame at the end of the LiDAR scan, as illustrated in Fig.~\ref{fig:ch6_propagation}. This backward propagation provides several advantages to the system: (1) given the relatively lower accuracy of magnetic-based localization, aggregating multiple observations enables a more robust update by solving a weighted least squares problem; (2) multiple observations aligned at the scan end time allow the estimation uncertainty to be quantified, enabling adaptive adjustment of the measurement noise during the filter update; and (3) LiDAR point undistortionrelies solely on IMU which ensure the quality. The magnetic-field based localization is introduced as follows:
\paragraph{AMF Mapping} Our magnetometer sensor suite consists of a tri-axis magnetometer array, in which magnetometers' readings at time $t_k$ will be a sequence of magnetic measurements $\mathcal{B}_{t_k} = \{\mathbf{B}_{t_k}^j\}_{j=1}^N$ with $\mathbf{B}_{t_k}^j = [B_{x, t_k}^j, B_{y, t_k}^j, B_{z, t_k}^j]^\top$ which are the measurements of $j^{th}$ magnetometer and $N$ is the number of magnetometers contained in this array. The $N$ magnetometers are hardware-synchronized so there is no timing difference among them. For the magnetic-based localization, the goal is to estimate the robot state according to the matching of a given pre-built magnetic map $\mathcal{M}(\cdot)$, which is a dense grid map describing the magnetic value at each point within a pre-defined area. To obtain such a prebuilt dense grid map, we adopt our previous research in \cite{lyu2024s} about sliding Gaussian Process Regression (s-GPR). A brief recap is given here. Let $S = \{(\mathbf{p}_1,\mathbf{B}_1), (\mathbf{p}_2,\mathbf{B}_2),...,(\mathbf{p}_n,\mathbf{B}_n)\}$ denote the set of magnetic fingerprints collected, where $\mathbf{p} \in \mathbb{R}^{n \times 3}$ represents the corresponding three-dimensional positions where the fingerprint is sampled. GPR can then be used to infer the posterior distribution of the latent magnetic field function $f(\mathbf{p})$ conditioned on the training dataset $S$ with Eq.~\eqref{eq:gpr_eq} and Eq.~\eqref{eq:gpr_cov}. 
\vspace{-0.5em}
\begin{equation} 
	\label{eq:gpr_eq}
	\small{
	\begin{aligned}
		f(\mathbf{p}^j)&\sim\mathcal{GP}{({m(\mathbf{p}^j), k(\mathbf{p}_m, \mathbf{p}_n))}} \\
		\mathbf{B}^j &= f(\mathbf{p}^j) + \epsilon
	\end{aligned}
}
\vspace{-0.5em}
\end{equation}
and the covariance (kernel) function, which is expressed as:
\vspace{-0.5em}
\begin{equation}
	\label{eq:gpr_cov}
	\small{
	\operatorname*{cov}(\mathbf{p}_m, \mathbf{p}_n) = k(\mathbf{p}_m, \mathbf{p}_n) + {\sigma}_n^2{\delta}_{mn}
}
\vspace{-0.5em}
\end{equation}
Since conventional GPR exhibits high computational complexity, the s-GPR strategy proposed in \cite{lyu2024s} is employed to accelerate the magnetic map construction process by using sliding windows. After obtaining the dense magnetic grid map, magnetic field values at arbitrary query points are computed via bilinear interpolation on the grid, producing continuous spatial estimates. To use such a map, the magnetic map value at a given query position $\mathbf{p}$ within the mapped region is defined as
\vspace{-0.5em}
\begin{equation}
	\small{
	\begin{aligned}
		\mathcal{P} &= \{\mathbf{p} \mid \mathbf{p} \in \mathbb{R}^3 \ \text{and within the mapped region}\}, \\
		\mathcal{M} &: \mathcal{P} \subset \mathbb{R}^3 \rightarrow \mathbb{R}^3, \qquad
		\mathbf{B}_{\text{map}} = \mathcal{M}(\mathbf{p})
	\end{aligned}
}
\vspace{-0.5em}
\end{equation}

\paragraph{Synchronization} In practice, the sampling times of the magnetometer and the IMU are generally asynchronous, which introduces additional motion distortion between the IMU sampling time $t_{i-1}$ and the magnetometer sampling time $t_{\gamma}$. Fortunately, since the full IMU states during the scan window have already been predicted, this motion distortion can be compensated using a formulation similar to those in Eq.~\ref{eq:forward_propagation} and Eq.~\ref{eq:backward_propagation}, yielding
${^{\mathcal{M}}\hat{\mathbf{T}}_{I_{t_\gamma}}}  = ( {^{\mathcal{M}}\hat{\mathbf{R}}_{I_{t_\gamma}}}, {^{\mathcal{M}}\hat{\mathbf{p}}_{I_{t_\gamma}}})$ which provides the IMU prediction at time $t_\gamma$ and transformation to the scan end time ${^{I_{t_{k}}}\check{\mathbf{T}}_{I_{t_\gamma}}}  = ( {^{I_{t_{k}}}\check{\mathbf{R}}_{I_{t_\gamma}}}, {^{I_{t_{k}}}\check{\mathbf{p}}_{I_{t_\gamma}}})$. 

\paragraph{AMF-based Localization} To incorporate magnetometer observations into the system, a two-step strategy is adopted. (1) First, purely magnetometer-based localization is performed, with the initial prediction provided by backward propagation. (2) The resulting poses are then propagated to the scan end time using backward propagation and subsequently fused into the system through the IESKF update. 

To achieve magnetic-based localization, given the magnetometer measurement at time $t_\gamma$, it can be expressed as follows:
\vspace{-0.5em}
\begin{equation}
	\small{
	[\hat{\mathbf{B}}_{\text{meas}}]_{t_\gamma}^j
	= {\mathbf{R}_{M_{t_\gamma}}}^{\top}\mathcal{M}(\mathbf{p}_{M_{t_\gamma}}^j) + \mathbf{n}_{t_\gamma}^j
}
\vspace{-0.5em}
\end{equation}
where $[\hat{\mathbf{B}}_{\text{meas}}]_{t_\gamma}^j$ is the magnetometer readings, $(\cdot)^j$ means the $j$-th magnetometers on an array, and $\mathbf{n}_{t_\gamma}^j$ is the measurement noise. Then, the localization problem is to find the optimal state $\mathbf{R}_{M_{t_\gamma}}$ and $\mathbf{p}_{M_{t_\gamma}}^j$ to minimize the cost function
\vspace{-0.5em}
\begin{equation}
	\label{eq:mag_loc_estimate}
	\small{
	\begin{aligned}
		\mathbf{x}_{t_\gamma}^{*}
		&:= \argmin_{\mathbf{x}_{t_\gamma}} \sum_{j=1}^{N} \left\| \mathbf{r}^j(\mathbf{x}_{t_\gamma}) \right\|^2 \\
		&= \argmin_{\mathbf{x}_{t_\gamma}}
		\sum_{j=1}^{N}
		\left\|
		[\hat{\mathbf{B}}_{\text{meas}}]_{t_\gamma}^j - {\mathbf{R}_{M_{t_\gamma}}^{\top}} \mathcal{M}(\mathbf{p}_{M_{t_\gamma}}^j)
		\right\|^2
	\end{aligned}
}
\vspace{-0.5em}
\end{equation}
To find the optimal solution, we apply the Gauss-Newton optimization.
The optimization is offered with the initial value given at the IMU state prediction ${^{\mathcal{M}}\hat{\mathbf{T}}_{I_{t_\gamma}}}  = ( {^{\mathcal{M}}\hat{\mathbf{R}}_{I_{t_\gamma}}}, {^{\mathcal{M}}\hat{\mathbf{p}}_{I_{t_\gamma}}})$ and extrinsic parameters between the magnetometers and the IMU $^{I}\mathbf{T}_{M}^j = (^{I}\mathbf{R}_M^j, ^{I}\mathbf{p}_M^j)$ yielding ${^{\mathcal{M}}\hat{\mathbf{T}}_{M_{t_\gamma}}}  = ( {^{\mathcal{M}}\hat{\mathbf{R}}_{M_{t_\gamma}}}, {^{\mathcal{M}}\hat{\mathbf{p}}_{M_{t_\gamma}}}) = ({^{\mathcal{M}}\hat{\mathbf{R}}_{I_{t_\gamma}}}^{I}\mathbf{R}_M^j, {^{\mathcal{M}}\hat{\mathbf{R}}_{I_{t_\gamma}}}^{I}\mathbf{p}_M^j+{^{\mathcal{M}}\hat{\mathbf{p}}_{I_{t_\gamma}}})$. Define the small update on rotation and translation as $\Delta\mathbf{x}=[\Delta\boldsymbol{\phi}^{\top}\, \Delta\mathbf{p}^\top\,\mathbf{0}]$. The new residual after small update:
\vspace{-0.5em}
\begin{equation}
	\small{
	\mathbf{r}^j(\hat{\mathbf{x}}_{t_\gamma} \boxplus \Delta\mathbf{x})
	=
	[\hat{\mathbf{B}}_{\text{meas}}]_{t_\gamma}^j - \left(\hat{\mathbf{R}}_{M_{t_\gamma}}\exp\!\left(\Delta\boldsymbol{\phi}\right)\right)^{\top}\mathcal{M}(\hat{\mathbf{p}}_{M_{t_\gamma}}^j + \Delta\mathbf{p})
}
\vspace{-0.5em}
\end{equation}
Taking the first-order Taylor expansion approximation is $\mathbf{r}^j(\hat{\mathbf{x}}_{t_\gamma} \boxplus \Delta\mathbf{x})
\approx
\mathbf{r}^j(\hat{\mathbf{x}}_{t_\gamma})
+
\mathbf{J}^j \Delta\mathbf{x}$. 
Here, $\mathbf{J}^j$ denotes the Jacobian matrix of the $j$-th magnetometer observation with respect to the state perturbation. It is given by
\vspace{-0.5em}
\begin{equation}
	\small{
	\begin{aligned}
		\mathbf{J}^j
		&=
		-
		\left[
		{\hat{\mathbf{R}}_{M_{t_{\gamma}}}}^{\top}
		\nabla \mathcal{M}(\hat{\mathbf{p}}_{M_{t_{\gamma}}}^{j})
		\;\middle|\;
		\big[
		{\hat{\mathbf{R}}_{M_{t_{\gamma}}}}^{\top}
		\mathcal{M}(\hat{\mathbf{p}}_{M_{t_{\gamma}}}^{j})
		\big]_{\times}
		\right]
	\end{aligned}
}
\vspace{-0.5em}
\end{equation}
where $\nabla \mathcal{M}(\hat{\mathbf{p}}_{M_{t_{\gamma}}}^{j})$ denotes the spatial gradient of the magnetic grid map evaluated at the current position, and $[\cdot]_{\times}$ represents the skew-symmetric matrix operator.
The Gauss--Newton state update is then obtained as $\hat{\mathbf{x}}_{t_\gamma}\boxplus\Delta\mathbf{x}$
with the perturbation $\Delta\mathbf{x}$ computed by
\vspace{-0.5em}
\begin{equation}
	\small{
	\Delta\mathbf{x}
	=
	-
	(
	\sum_{j=1}^{N}
	{\mathbf{J}^j}^{\top} \mathbf{J}^j
	)^{-1}
	\sum_{j=1}^{N}
	{\mathbf{J}^j}^{\top} \mathbf{r}^j
}
\vspace{-0.5em}
\end{equation}
Note that the localization output using the magnetometer array's measurements is in the global map frame $\mathcal{M}$. Thus, the transformed measured localization of IMU body pose at time $t_k$ can be derived as ${^{\mathcal{M}}\mathbf{p}_{I_{t_k}}} = {^{\mathcal{M}}\hat{\mathbf{T}}_{I_{t_k}}}{^{I_{t_k}}\check{\mathbf{T}}_{I_{t_\gamma}}}{^{\mathcal{M}}\hat{\mathbf{T}}_{I_{t_\gamma}}^{\top}}{^{\mathcal{M}}\mathbf{T}_{M_{t_\gamma}}}{^{M}\mathbf{p}_{I}}$ and $	{^{\mathcal{M}}\mathbf{R}_{I_{t_k}}} = {^{\mathcal{M}}\hat{\mathbf{R}}_{I_{t_k}}}{^{I_{t_k}}\check{\mathbf{R}}_{I_{t_\gamma}}}{^{\mathcal{M}}\hat{\mathbf{R}}_{I_{t_\gamma}}^{\top}}{^{\mathcal{M}}\mathbf{R}_{M_{t_\gamma}}}{^{I}\mathbf{R}_M^{\top}}$
where $^{I}\mathbf{R}_M$ and $^{M}\mathbf{p}_I$ are the extrinsic parameters between the magnetometer array and the IMU. This transformed observation eventually provides a direct measurement over the state ${^{\mathcal{M}}\mathbf{p}_I}$ and ${^{\mathcal{M}}\mathbf{R}_I}$.
\vspace{-1em}
\subsection{State Update}
Based on Eq.~\ref{eq:mag_loc_estimate}, the system state can be updated using a sequence of magnetic observations $\{({^{\mathcal{M}}\mathbf{R}_n}, {^{\mathcal{M}}\mathbf{p}_n}) \mid n = 1, 2, \ldots, L\}$, where $L$ denotes the number of magnetometer measurements received within a LiDAR scan window. Since the observations are directly defined on the state space, a similar IESKF formulation as in the LiDAR update can be adopted. The residual at iteration $\kappa$ is then defined as
\vspace{-0.5em}
\begin{equation}
	\small{
	\mathbf{r}_n^\kappa(\hat{\mathbf{x}}_{t_k}) = 
	\begin{bmatrix}
		\mathbf{r}_{\mathbf{p}_{n}}^\kappa(\hat{\mathbf{x}}_{t_k}) \\
		\mathbf{r}_{\mathbf{\boldsymbol{\theta}}_{n}}^\kappa(\hat{\mathbf{x}}_{t_k})
	\end{bmatrix} = 
	\begin{bmatrix}
		{^{\mathcal{M}}\mathbf{p}_n} - {^{\mathcal{M}}\hat{\mathbf{p}}_{I_{t_k}}} \\ \log\left({^{\mathcal{M}}\hat{\mathbf{R}}_{I_{t_k}}^{\top}}{^{\mathcal{M}}\mathbf{R}_n}\right)
	\end{bmatrix}\in{\mathbb{R}^6}
}
\vspace{-0.5em}
\end{equation}
Adding the right perturbation $\delta\mathbf{x}$ to the residual and taking the first order approximation results in
\vspace{-0.5em}
\begin{equation}
	\small{
	\begin{aligned}
	&\approx\mathbf{r}_n^\kappa(\hat{\mathbf{x}}_{t_k}^\kappa) + {^{M}\mathbf{H}_n^\kappa}\delta\mathbf{x}
	\end{aligned}
}
\vspace{-0.5em}
\end{equation}
where $\mathbf{H}_n^\kappa\in\mathbb{R}^{6\times6}$ is the partial derivative with respect to $\delta\mathbf{x}^\kappa$:
\vspace{-0.5em}
\begin{equation}
	\small{
	{^{M}\mathbf{H}_n^\kappa} = 
	\begin{bmatrix}
		-\mathbf{I}_3 & \mathbf{0} \\
		\mathbf{0} & -\mathbf{J}^{-1}(	\mathbf{r}_{\mathbf{\boldsymbol{\theta}}_{n}}^\kappa)
	\end{bmatrix}\approx
	-\begin{bmatrix}
		\mathbf{I}_3 & \mathbf{0} \\
		\mathbf{0} & \mathbf{I}_3
	\end{bmatrix}
}
\vspace{-0.5em}
\end{equation}
if the residual is small. Then, the maximum a-posteriori estimate (MAP) becomes
\vspace{-0.5em}
\begin{equation}
	\small{
	\label{eq:map_mag}
	\argmin_{\delta\mathbf{x}}\left[\frac{1}{2}\|\hat{\mathbf{x}}_{t_k}^\kappa\boxminus\hat{\mathbf{x}}_{t_k}\|_{\hat{\mathbf{P}}_{t_k}^{-1}}^2+\frac{1}{2}\sum_{n=1}^{L}\|\mathbf{r}_n^\kappa(\hat{\mathbf{x}}_{t_k}^\kappa) + {^{M}\mathbf{H}_n^\kappa}\delta\mathbf{x}\|_{\mathbf{R}_{n}^{-1}}^2\right]
}
\vspace{-0.5em}
\end{equation}
where $\|\mathbf{x}\|_{\mathbf{A}} = \mathbf{x}^{\top}\mathbf{A}\mathbf{x}$. It can be observed that the contribution of each observation is governed by the observation noise covariance matrix $\mathbf{R}$. Therefore, by appropriately designing $\mathbf{R}$, the confidence assigned to each observation can be adaptively adjusted. The subsequent steps follow the standard iterated Kalman filter gain computation and state update procedure.
Suppose ${^{M}\mathbf{H}_{t_k}^\kappa} = [{^{M}{\mathbf{H}_1^\kappa}^{\top}}, {^{M}{\mathbf{H}_2^\kappa}^{\top}}, ..., {^{M}{\mathbf{H}_m^\kappa}^{\top}]^{\top}}$, $\mathbf{R} = \mathrm{diag}(\mathbf{R}_1, \mathbf{R}_2, ..., \mathbf{R}_n)$, and $\mathbf{r}_{t_k}^\kappa = [{\mathbf{r}_1^\kappa}^{\top}, {\mathbf{r}_2^\kappa}^{\top}, ..., {\mathbf{r}_n^\kappa}^{\top}]$, then
\vspace{-0.5em}
\begin{equation}
	\label{eq:ieskf_update}
	\small{
	\begin{aligned}
		\mathbf{K}_{t_k}^\kappa &= \hat{\mathbf{P}}_{t_k}{\mathbf{H}_{t_k}^\kappa}^{\top}{\mathbf{S_{t_k}^\kappa}}^{-1} \\
		\delta\mathbf{x}^* &= -\mathbf{K}_{t_k}^\kappa\mathbf{r}_{t_k}^\kappa - (\mathbf{I} - \mathbf{K}_{t_k}^\kappa{{^{M}\mathbf{H}_{t_k}^\kappa}})(\hat{\mathbf{x}}_{t_k}^{\kappa} \boxminus \hat{\mathbf{x}}_{t_k}) \\
		\hat{\mathbf{x}}_{t_k}^{\kappa + 1} &= \hat{\mathbf{x}}_{t_k}^{\kappa}\boxplus\delta\mathbf{x}^*
	\end{aligned}
}
\vspace{-0.5em}
\end{equation}
where $	\mathbf{S_{t_k}^\kappa} = \mathbf{H}_{t_k}^\kappa\hat{\mathbf{P}}_{t_k}{\mathbf{H}_{t_k}^\kappa}^{\top}+\mathbf{R}$.
Eventually, if the IESKF successfully converges, the optimal estimation will be:
\vspace{-0.5em}
\begin{equation}
	\small{
	\bar{\mathbf{x}}_{t_k} = \hat{\mathbf{x}}_{t_k}^{\kappa + 1}
}
\vspace{-0.5em}
\end{equation}
Then, the updated state will be used as the initial value for LiDAR scan-to-map matching before the LiDAR update.
\vspace{-1em}

\subsection{Degeneration Detection and Outlier Filtering}
It is important to analyze the degeneracy and reliability conditions of each sensing modality in the fused system to achieve optimal performance. Specifically, the filter should assign higher confidence to reliable observations while rejecting outliers. Therefore, both LiDAR and magnetometer measurements are evaluated in the proposed fusion framework.

\begin{figure*}[!t]
	\centering
	\includegraphics[width=0.99\linewidth]{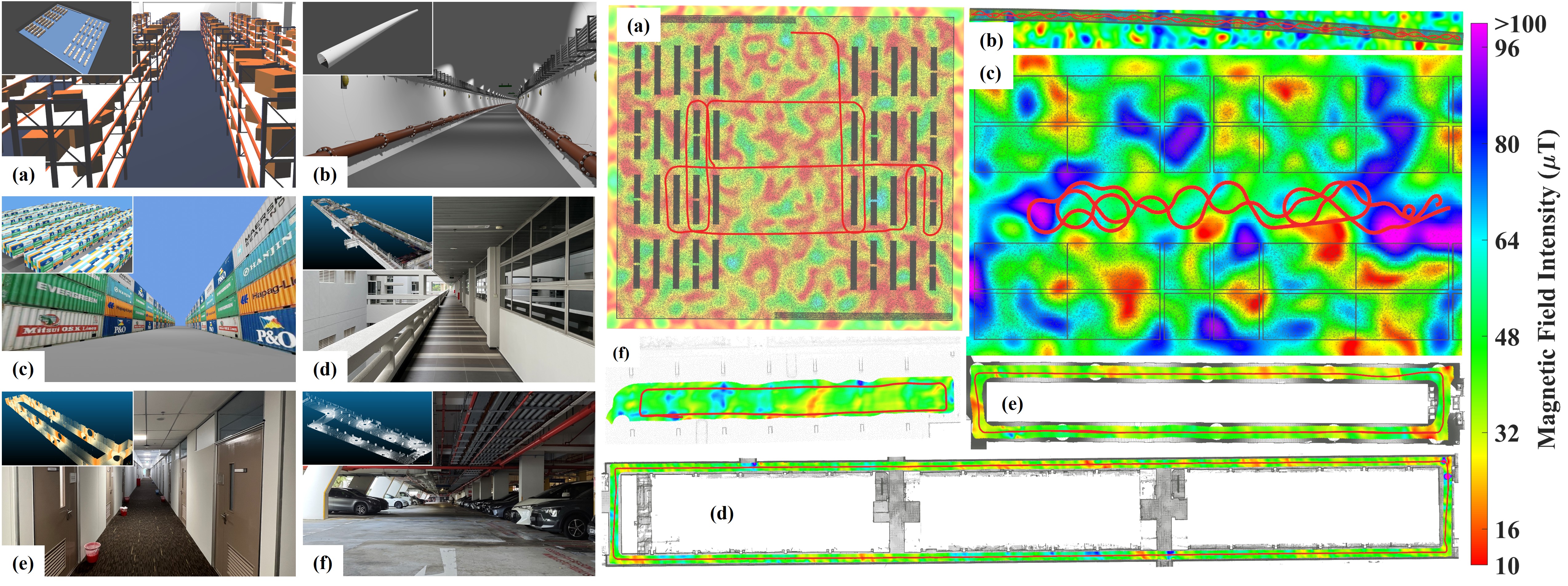}
	\caption{Left side: Visualization of different experimental environments, including simulated (a) industrial warehouse, (b) vehicle tunnel, (c) logistics seaport, and real-world (d) semi-indoor walkway, (e) indoor office corridor, and (f) vehicle carpark. The top-left corner presents an overview of each environment. Right side: Examples of the trajectory used for the experiments in different scenarios and the visualization of the magnetic field intensity at different positions. }
	\label{fig:exp_traj}
	\vspace{-1em}
\end{figure*}
\paragraph{LiDAR Degeneration Detection} For LiDAR, degeneracy occurs when certain state directions are insufficiently constrained by the observed data. This is analyzed through the Hessian matrix, following the approach of Zhang et al.~\cite{zhang2016degeneracy}. The key difference in this work is that the projection is performed directly on the Hessian matrix rather than on the solution $\delta \mathbf{x}$, since the estimated solution under degeneracy may already contain amplified noise from the measurements~\cite{tuna2025informed}. By projecting the Hessian onto well-constrained directions, the influence of degenerate components can be effectively mitigated. This is achieved by performing eigenvalue decomposition on the Hessian matrix,
\vspace{-0.5em}
\begin{equation}
	\small{
	{^{L}\mathbf{H}^{\top}}\mathbf{R}^{-1}{^{L}\mathbf{H}} = \mathbf{U}\boldsymbol{\Sigma}\mathbf{V}^{\top}
}
\vspace{-0.5em}
\end{equation}
where ${^{L}\mathbf{H}} = [{^{L}{\mathbf{H}_1^\kappa}^{\top}}, {^{L}{\mathbf{H}_2^\kappa}^{\top}}, ..., {^{L}{\mathbf{H}_m^\kappa}^{\top}]^{\top}}$ is the Hessian matrix which is constructed from LiDAR features, where each row corresponds to a feature point in the LiDAR scan. The degeneracy analysis is performed only on the leftmost six columns of $\mathbf{H}$, which correspond to the pose states. The diagonal matrix $\boldsymbol{\Sigma}$ contains the eigenvalues, \textit{i.e.}, $\mathrm{diag}(\sigma_1, \sigma_2, \ldots, \sigma_6)$. In degenerate directions, the corresponding eigenvalues tend to be small. Therefore, these small eigenvalues are truncated, and the Hessian matrix is reprojected accordingly:
\vspace{-0.5em}
\begin{equation}
	\small{
	d_i = \begin{cases}
		1, &\text{if } \sigma_i > \tau \\
		0, &\text{if } \sigma_i < \tau
	\end{cases}
}
\vspace{-0.5em}
\end{equation}
where $\tau$ is the threshold for degenerate eigenvalue, which was chosen as 15 in this case, and $\mathbf{D} = \mathrm{diag}(d_1, d_2, ..., d_6)$. Then the reprojected Hessian matrix ${^{L}\mathbf{H}_g}$ is equivalent to 
\vspace{-0.5em}
\begin{equation}
	\small{
	{^{L}\mathbf{H}_g} = {^{L}\mathbf{H}}\mathbf{V}\mathbf{D}\mathbf{V}^{\top}
}
\vspace{-0.5em}
\end{equation}

\begin{figure}[!t]
	\centering
	\resizebox{0.99\linewidth}{!}{%
		\begin{minipage}{\linewidth}
			\centering
			\subfloat[Sensor Suite\label{fig:sensor_suite}]{%
				\includegraphics[width=0.325\linewidth]{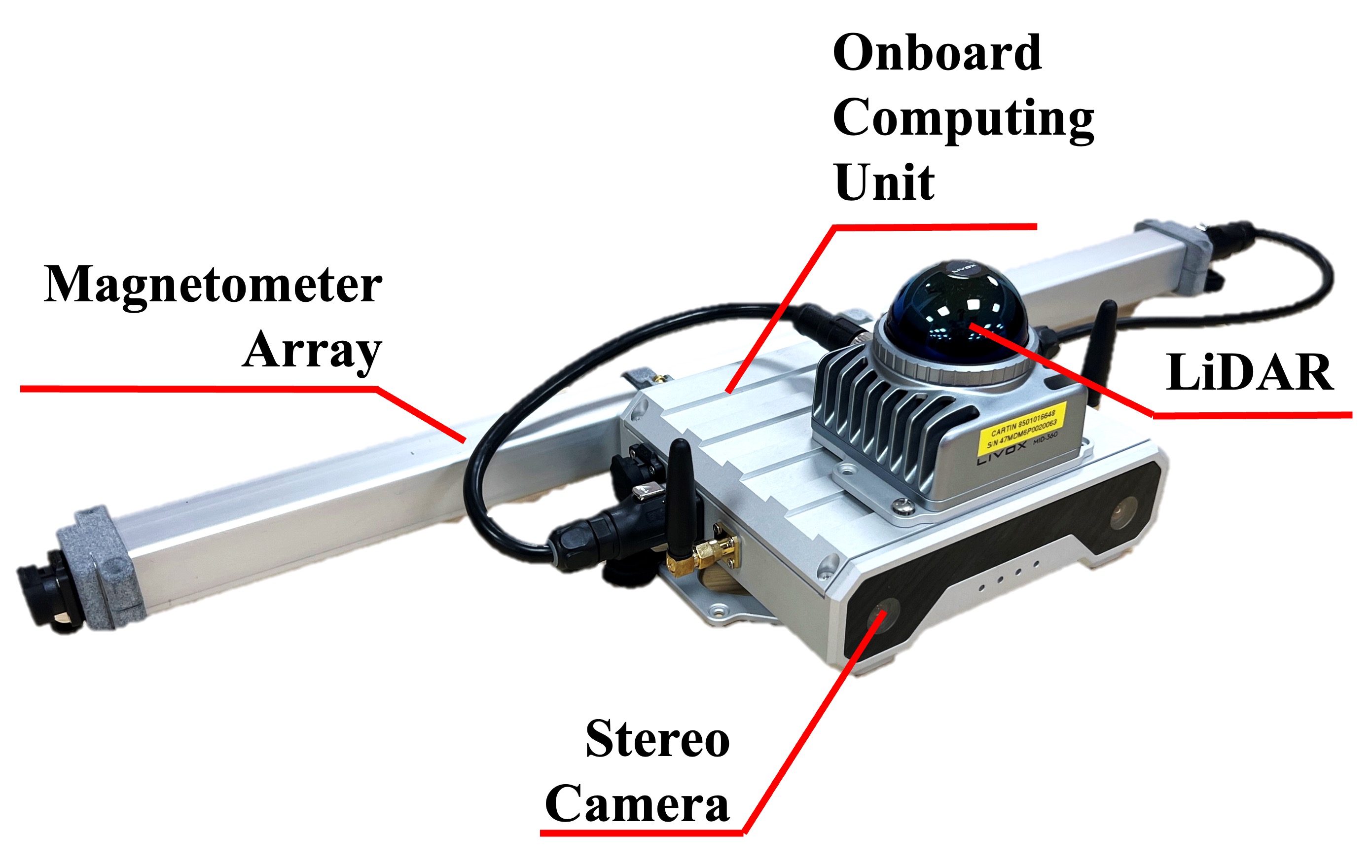}}
			\hfill
			\subfloat[Real Scout Mini\label{fig:ch6_real_scout}]{%
				\includegraphics[width=0.325\linewidth]{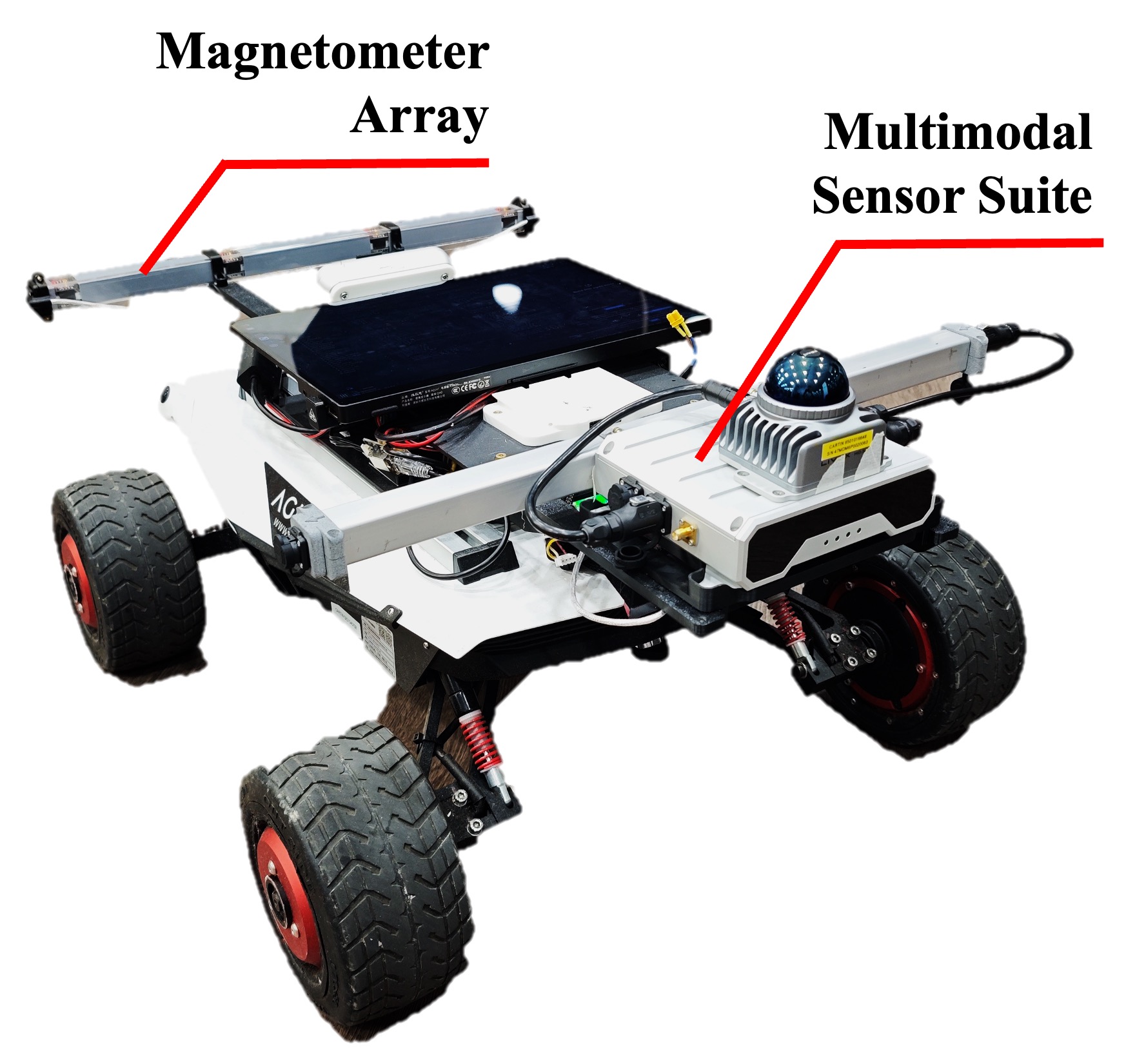}}
			\hfill
			\subfloat[Simulated Scout Mini\label{fig:ch6_simu_scout}]{%
				\includegraphics[width=0.325\linewidth]{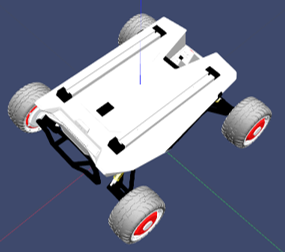}}
		\end{minipage}
	}
	\caption{The Scout Mini AMR is employed in both real-world and simulated environments to collect experimental data.}
	\vspace{-1.5em}
	\label{fig:ch6_amr_platform}
\end{figure}
\paragraph{Magnetometer Outlier Rejection} For the magnetometer, a different strategy is adopted. Due to the relatively lower accuracy of magnetic-based localization compared with LiDAR (\textit{e.g.}, approximately 10\,cm versus 3\,cm, as reported in \cite{lyu2026roslacrobustsimultaneouslocalization}), the proposed multimodal system is designed to rely on LiDAR for accuracy when it is reliable, while using magnetic information to constrain the degenerate directions of LiDAR. To achieve this, two strategies are employed: 1) when no LiDAR degeneracy is detected, the magnetometer residual $\mathbf{r}_{t_k}$ is manually scaled down to prevent magnetic observations from dominating the update $\tilde{\mathbf{r}}_{t_k} = \beta\mathbf{r}_{t_k}$
where $\beta$ denotes the shrinkage factor. This scaling is applied instead of inflating the measurement covariance to avoid unintended inflation of $\mathbf{P}_{t_k}$ prior to the LiDAR-based update; 2) when LiDAR degeneracy is detected in certain directions, the Hessian matrix associated with the magnetic residual is inversely reprojected to enhance constraints along those degenerate directions:
\vspace{-0.5em}
\begin{equation}
	\vspace{-0.5em}
	\small{
	{^{M}\mathbf{H}_u} = {^{M}\mathbf{H}}\mathbf{V}(\mathbf{I} - \mathbf{D})\mathbf{V}^{\top}
	}
	\vspace{-0.3em}
\end{equation}
where the reprojected ${^{M}\mathbf{H}_u}$ can be put into Eq.~\ref{eq:map_mag} and Eq.~\ref{eq:ieskf_update} for estimation. 

To further ensure the reliability of magnetic-based localization, outliers corresponding to unreliable magnetic pose estimates are removed from the observations. To achieve this, a robust kernel is applied to the magnetic observations prior to the update. Consistent with the IESKF framework, the robust kernel is defined on the Mahalanobis distance $\mathbf{s}_{t_k} = [s_1, s_2, ..., s_n]$ with $s_n = \sqrt{{\mathbf{r}_{t_k}}^{\top}\mathbf{S}_{t_k}\mathbf{r}_{t_k}}$
where $\mathbf{r}_{t_k}$ is the residual between the magnetic-based localization result $\mathbf{x}_{t_k}$ and the predicted pose from IMU forward propagation $\hat{\mathbf{x}}_{t_k}$, and $\mathbf{S}_{t_k}$ is the innovation covariance. Applying the Huber kernel to the Mahalanobis distance results in
\vspace{-0.5em}
\begin{equation}
	\small{
	w =
	\begin{cases}
		1, &\text{if } s \leqslant c \\
		\frac{c}{s}, &\text{if } s > c
	\end{cases}
}
\vspace{-0.5em}
\end{equation}
where $c$ is the threshold, and is chosen to be 3.55 in this case. Suppose
\vspace{-0.5em}
\begin{equation}
	\small{
	\mathbf{W}_{t_k} = 
	\begin{bmatrix}
		w_1\mathbf{I} & 0 & \dots & 0 \\
		0 & w_2\mathbf{I} & \dots & 0 \\
		\vdots & \vdots & \ddots & \vdots \\
		0 & 0 & \dots &  w_n\mathbf{I}
	\end{bmatrix}
}
\vspace{-0.5em}
\end{equation}
Finally,
\vspace{-0.5em}
\begin{equation}
	\label{eq:inflated_R}
	\small{
	\tilde{\mathbf{R}}_{t_k} = \mathbf{W}_{t_k}^{-1}\mathbf{R}
}
\vspace{-0.5em}
\end{equation}
where Eq.~\ref{eq:inflated_R} can be substituted back to Eq.~\ref{eq:ieskf_update} as an adaptive observation noise in the Kalman gain calculation.

\section{Experiments and Analysis}\label{sec:chapter6/exp}

\subsection{Evaluation Overview}
In this section, multiple experiments are conducted in both high-fidelity simulated environments and real-world scenarios to evaluate the performance of the proposed method. In all experiments, the algorithms are executed on the same desktop computer equipped with an Intel i7 processor at 2.3GHz and 64GB of RAM. An autonomous robot with an identical configuration is deployed in both simulated and real-world environments, ensuring consistency of experimental results across different scenarios. In the simulation environments, ground-truth odometry is directly available.

The details of the experimental environments are as follows:
\begin{itemize}
	\item \textbf{High-fidelity Simulated Environments:} The Gazebo\footnote{\url{https://gazebosim.org/home}} simulation platform is employed to emulate multiple realistic environments, including (a) an industrial warehouse, (b) a vehicle tunnel, and (c) a logistics seaport. As shown in Fig.~\ref{fig:exp_traj}, the industrial warehouse has a size of $90\,\mathrm{m} \times 80\,\mathrm{m}$ and contains repetitive storage racks, reinforced concrete pillars, and conveyor belts, with featureless open spaces between the racks. The logistics seaport spans $300\,\mathrm{m} \times 300\,\mathrm{m}$ and consists of multiple container yards, with traffic lanes for AMRs of width $18\,\mathrm{m}$ between them. The tunnel environment simulates a realistic featureless tunnel of length $200\,\mathrm{m}$ with slight curvature. In all environments, a simulated Scout Mini mobile robot is deployed to collect data, as shown in Fig.~\ref{fig:ch6_simu_scout}, equipped with one Ouster OS-32 LiDAR and a total of eight idealized magnetometers mounted at the front and rear of the robot chassis. The magnetic measurements are corrupted with zero-mean Gaussian noise with a standard deviation of $\sigma = 1\,\mu\mathrm{T}$.
	
	\item \textbf{Real-world Environments:} Three representative real-world environments, including (d) a semi-indoor walkway, (e) an indoor office corridor, and (f) a vehicle carpark, are evaluated using both geometric information from LiDAR and magnetic information from magnetometers collected by a real Scout Mini AMR platform, as shown in Fig.~\ref{fig:ch6_real_scout}. As illustrated in Fig.~\ref{fig:exp_traj}, the semi-indoor office walkway has a size of $181\,\mathrm{m} \times 23\,\mathrm{m}$, while the indoor corridor measures $15\,\mathrm{m} \times 65\,\mathrm{m}$. Both environments contain either repetitive geometric structures or featureless regions. The carpark, with a size of $56\,\mathrm{m} \times 7\,\mathrm{m}$, exhibits time-varying magnetic fields caused by the presence and movement of vehicles. The platform is equipped with our self-designed sensor suite (Fig.~\ref{fig:sensor_suite}) with a Mid-360 LiDAR and eight RM3100 magnetometers mounted on the AMR chassis. All sensors are calibrated both intrinsically and extrinsically. The magnetic map is constructed with the assistance of a modified version of CTE-MLO~\cite{cte_mlo} based on a Leica total-station-based pre-built map, and the two maps are aligned to the same global frame.
\end{itemize}
\vspace{-1em}
\subsection{Evaluation Protocol}

\subsubsection{Comparison Baseline}
Quantitative analysis is conducted to evaluate the robustness and effectiveness of the proposed method compared with state-of-the-art approaches for localization in various degenerate environments. In LiDAR-degenerate scenarios, the compared methods include: (1) HDL Localization~\cite{koide2019portable}, an NDT-based LiDAR-inertial localization method; (2) CTE-MLO~\cite{cte_mlo}, a continuous-time LiDAR-only localization approach with Gaussian process-based trajectory representation; and (3) FAST-LIO2~\cite{fast_lio2}, a robust tightly coupled LiDAR-inertial system based on point-to-plane matching, modified to support localization on a pre-built map\footnote{\url{https://github.com/iDonghq/FAST_LIO_LOCALIZATION_PLUS}}. All methods are evaluated under their default configurations, with a maximum LiDAR sensing range limited to $100\,\mathrm{m}$.

In scenarios with magnetic map disturbances, the comparison includes: (1) RoSLAC, a robust magnetic-based localization method with online calibration capability as introduced in \cite{lyu2026roslacrobustsimultaneouslocalization}; (2) IDF-MFL~\cite{idf_mfl}, a stochastic optimization and sampling-based magnetic localization approach; and (3) PF~\cite{shi2022PDR}, a Monte Carlo sampling-based magnetic localization method. 

\subsubsection{Evaluation Metrics}
The localization accuracy is evaluated differently in simulated and real-world environments. In simulated environments, the ground-truth trajectory is directly obtained from the simulator. In real-world environments, the reference trajectory is generated using a modified version of CTE-MLO~\cite{cte_mlo}, which performs relocalization on a Leica total-station-based pre-built point cloud map. The localization performance is evaluated using the Absolute Trajectory Error (ATE). A lower ATE generally indicates better localization performance.
\vspace{-0.5em}

\subsection{Accuracy Evaluation}
\begin{table}[]
	\caption{Localization Absolute Trajectory Error (ATE) of multiple methods under different test environments. The failed estimation (pose estimated beyond map boundary) is marked as $/$, and the best results are marked \textbf{BOLD}}
	\label{tab:loc_ate}
	\renewcommand{\arraystretch}{1.3}
	\begin{adjustbox}{width=0.99\linewidth}
		\begin{tabular}{lc|cccccc}
			\hline\hline
			Scenarios                   & Seq & HDL    & CTE-MLO & FAST-LIO2       & \textbf{Proposed} & FAST-LIO2-10 & Proposed-10 \\ \hline
			\multirow{2}{*}{Warehouse} & 1   & 0.038  & 0.036   & \textbf{0.024} & 0.024             & /           & 0.024       \\
			& 2   & 0.036  & 0.040    & \textbf{0.025} & 0.025             & /           & 0.030        \\ \hline
			\multirow{2}{*}{Tunnel}    & 3   & /      & 61.991  & /              & \textbf{0.029}    & /           & 0.029       \\
			& 4   & 59.594 & 62.822  & 59.694         & \textbf{0.040}     & /           & 0.040        \\ \hline
			Seaport                    & 5   & 58.207 & /       & 0.053          & \textbf{0.052}    & /           & 0.056       \\ \hline
			\multirow{2}{*}{Walkway}   & 6   & /      & /       & \textbf{0.064} & 0.067             & 0.063       & 0.067       \\
			& 7   & /      & /       & \textbf{0.121} & 0.123             & 0.123       & 0.128       \\ \hline
			\multirow{2}{*}{Corridor}  & 8   & 0.246  & /       & \textbf{0.113} & 0.178             & /           & 0.192       \\
			& 9   & 0.248  & /       & \textbf{0.106} & 0.128             & /           & 0.141       \\ \hline\hline
		\end{tabular}
	\end{adjustbox}
	\vspace{-2em}
\end{table}

The localization robustness and accuracy across different scenarios are first evaluated. For each scenario, one or two sequences are collected using the Scout Mini platform, with trajectories of varying shapes or collection speeds, as shown in Fig.~\ref{fig:exp_traj}(a)-(f). The ATE with respect to the reference trajectories is reported in Table~\ref{tab:loc_ate}. The results indicate that all five environments are challenging, where both HDL Localization and CTE-MLO fail in most cases. FAST-LIO2 demonstrates higher robustness, with more success sequences, but still encounters difficulties in tunnel environments due to insufficient geometric features. In contrast, the proposed method, which incorporates magnetic-based localization, achieves 100\% localization success across all tests. In the $-10$ setting (last two columns of Table~\ref{tab:loc_ate}), where the LiDAR sensing range is restricted to $10\,\mathrm{m}$, leading to severe geometric degeneration, the proposed method maintains stable localization performance across all scenarios. This demonstrates the effectiveness of integrating magnetic information and highlights the improved robustness of the proposed multimodal system compared to single-modality approaches, particularly in environments with insufficient geometric features.

In terms of localization accuracy, the pure LiDAR-based method using FAST-LIO2 achieves the highest accuracy in non-degenerate scenarios ($0.024\,\mathrm{m}$--$0.121\,\mathrm{m}$), while the proposed fused system exhibits slightly lower accuracy ($0.024\,\mathrm{m}$--$0.178\,\mathrm{m}$). This behavior is expected, as magnetic-based localization generally has lower accuracy than LiDAR. In the proposed framework, when LiDAR degeneration is detected, magnetic observations contribute more significantly to the state estimation, which may lead to reduced overall accuracy. 

This observation is further supported by the results in the two $-10$ columns. Under this configuration, the system relies predominantly on magnetic-based localization due to severe LiDAR degeneration, which leads to a further reduction in accuracy. Nevertheless, even in this case, the system maintains an ATE of at most $0.192\,\mathrm{m}$.
More importantly, in scenarios where pure LiDAR-based methods completely fail, the proposed system continues to provide reliable pose estimation. Therefore, although the proposed multimodal fusion framework demonstrates a slight trade-off between localization accuracy and robustness, it significantly improves system reliability in challenging environments.
\vspace{-0.5em}

\subsection{Robustness Evaluation}
\begin{figure}[!t]
	\centering
	\resizebox{0.999\linewidth}{!}{%
		\begin{minipage}{\linewidth}
			\centering
			\subfloat[Well-constrained Environment\label{fig:well_constrained_ate}]{%
				\includegraphics[width=0.49\linewidth]{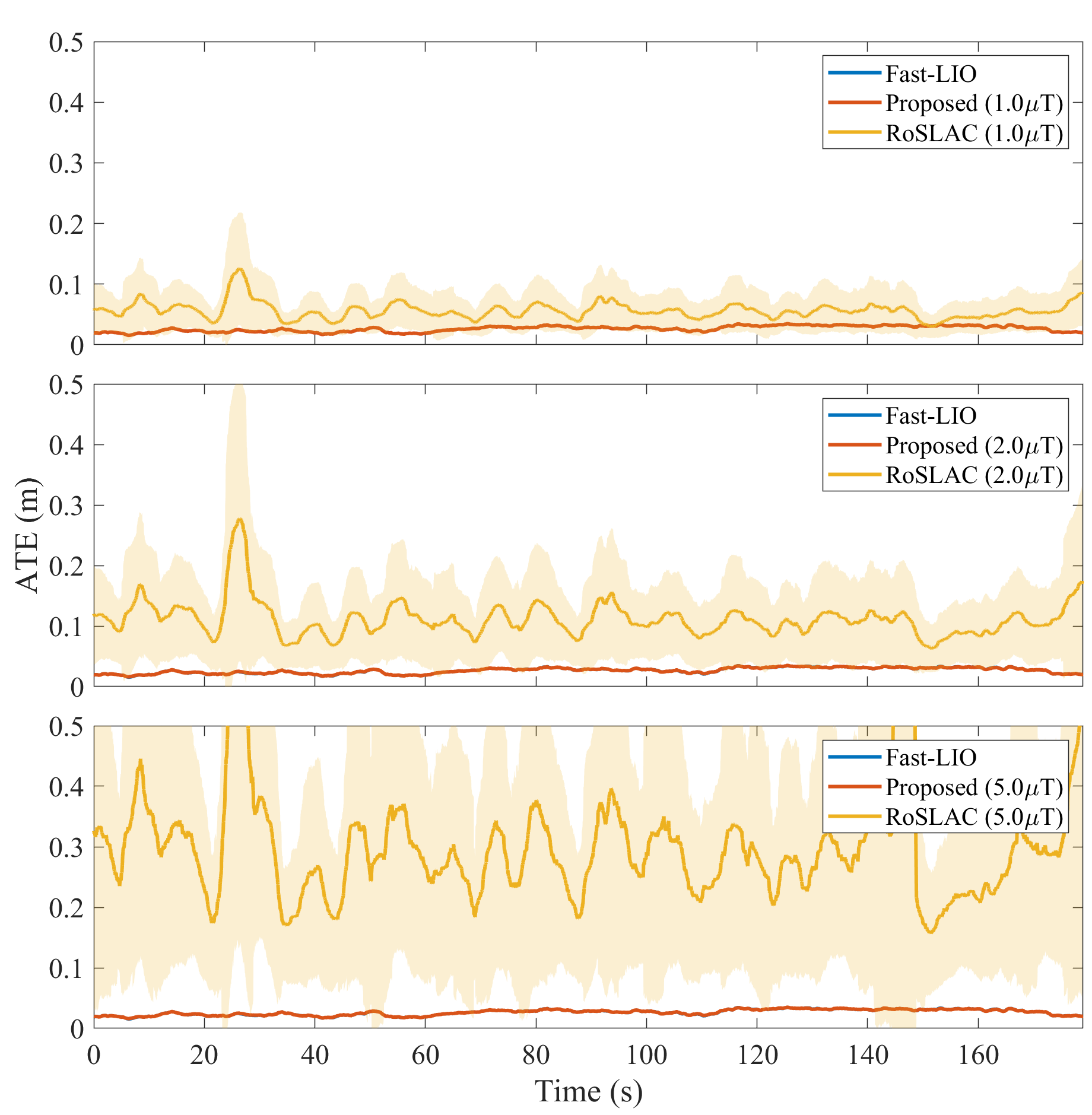}}
			\hfill
			\subfloat[LiDAR Degenerated Environment\label{fig:degenerate_ate}]{%
				\includegraphics[width=0.49\linewidth]{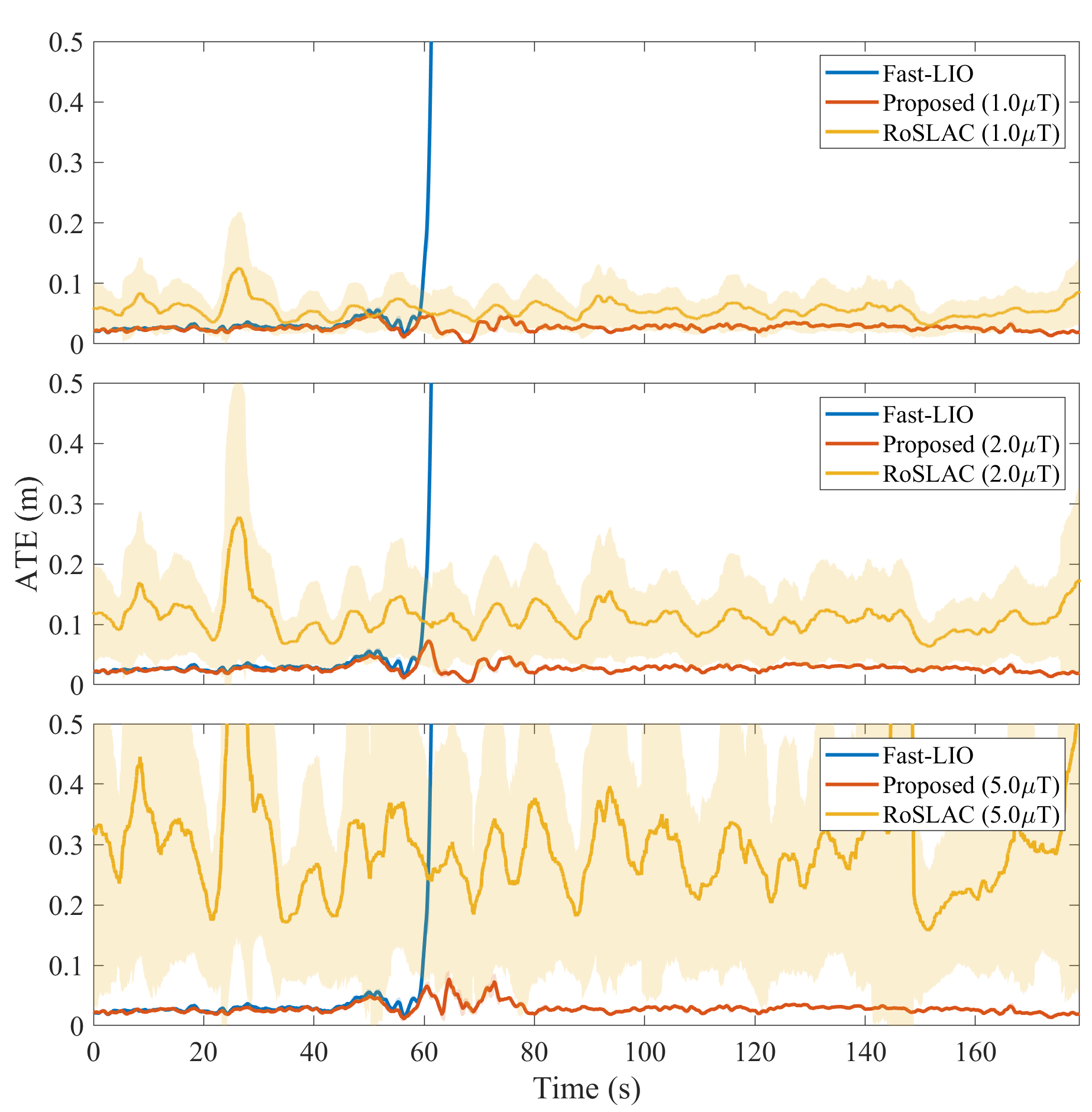}}
		\end{minipage}
	}
	\caption{Comparison of ATE (m) of Seq 1 in simulated warehouse environment of FAST-LIO2, RoSLAC, and the proposed MIL-LC method: (a) LiDAR estimation is well constrained, and (b) LiDAR estimation become degenerate. The confidence region represents the standard deviation within each window.}
	\label{fig:ate_comparison}
	\vspace{-1em}
\end{figure}

To further illustrate the robustness of the proposed method, the localization ATE of $\textit{Seq 6}$ is visualized over time in Fig.~\ref{fig:ate_comparison}, where the x-axis represents time and the y-axis represents ATE. In this comparison, purely LiDAR-based localization, purely magnetic-based localization (RoSLAC \cite{lyu2026roslacrobustsimultaneouslocalization}), and the proposed multimodal fusion method are evaluated. For a fair comparison, the LiDAR settings are kept identical across each column of subfigures, and the RoSLAC method is evaluated without any auxiliary odometry (\textit{e.g.}, wheel odometry or sequence accumulation). Without loss of generality, a constant-time window mean filter with a duration of $1\,\mathrm{s}$ is applied to the ATE of all methods for improved visualization. 
To ensure fairness, the same filtering is applied to all methods, and the standard deviation within each window is plotted as a confidence region, where a larger region indicates higher instability in the localization results.

The left column in Fig.~\ref{fig:well_constrained_ate} illustrates the scenario in which LiDAR does not exhibit degeneracy. It can be observed that, under this condition, the ATE of the proposed method almost overlaps with that of FAST-LIO2, indicating comparable performance. This is attributed to the degeneracy detection mechanism that when LiDAR measurements are reliable, the system suppresses magnetic-based updates and relies primarily on LiDAR for better accuracy. Consequently, although the purely magnetic-based localization exhibits significant jitter, the fused localization remains stable and accurate. For each row of the subfigure, different levels of additional Gaussian noise are injected into the magnetometer measurements to degrade their localization performance. It can be observed that the RoSLAC method, which relies solely on magnetic information, is significantly affected by a much higher mean error in the $5.0\,\mu\mathrm{T}$ noise scenario compared to the $1.0\,\mu\mathrm{T}$ case. In contrast, the proposed method shows negligible sensitivity to the magnetic noise level, as LiDAR dominates the pose estimation in this well-constrained scenario.

The right column in Fig.~\ref{fig:degenerate_ate} illustrates the scenario in which LiDAR becomes degenerate when the robot traverses a large featureless region between shelves (the LiDAR range is limited to $10\,\mathrm{m}$ to enforce this degeneration), departing from the left-side shelves at approximately $60\,\mathrm{s}$ and reaching the right-side shelves at around $80\,\mathrm{s}$. As expected, LiDAR degeneracy occurs, with a significant increase in ATE observed at $60\,\mathrm{s}$. In contrast, the proposed fused system maintains a low ATE, remaining below $0.1\,\mathrm{m}$ during this period. This is because magnetic-based localization dominates the pose estimation under LiDAR degeneration, although it introduces slightly increased jitter, and this jitter becomes more pronounced as additional noise is introduced into the magnetometer measurements. Notably, after $80\,\mathrm{s}$, when LiDAR becomes well-constrained again, the overall ATE rapidly decreases and stabilizes below $0.05\,\mathrm{m}$. This indicates that, upon re-entering a geometrically rich region, magnetic-based localization provides a reliable initial estimate for LiDAR-based localization, after which the system can successfully transition back to a LiDAR-dominant regime.
\begin{figure}[!t]
	\centering
	\includegraphics[width=0.999\linewidth]{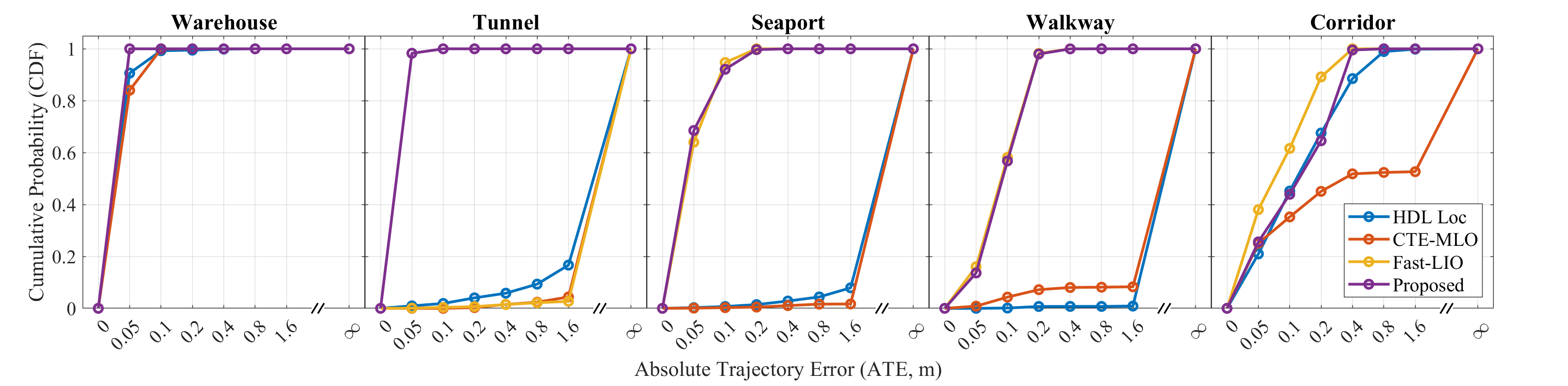}
	\caption{Cumulative probability of the Absolute Trajectory Error (ATE, m) of different localization methods in different scenarios. }
	\label{fig:loc_cdf}
	\vspace{-0.5em}
\end{figure}
\begin{figure}[!t]
	\centering
	\includegraphics[width=0.999\linewidth]{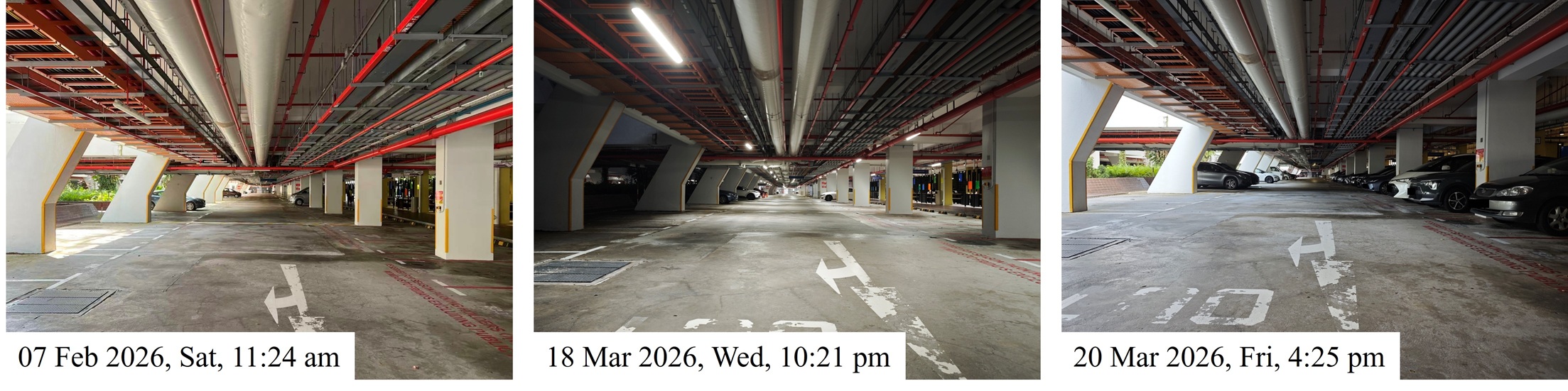}
	\caption{An illustration of the change of vehicles in the carpark environments. The pictures of the carpark are taken at three different timeslots, where it can be observed that sometimes there are more vehicles and sometimes the carpark is empty.}
	\label{fig:carpark_change}
	\vspace{-1em}
\end{figure}

Additional sequences are further visualized in Fig.~\ref{fig:loc_cdf} to evaluate robustness. In the cumulative distribution function (CDF) plot, the x-axis represents the localization ATE divided into multiple intervals, while the y-axis shows the proportion of localization results within each sequence that fall below a given error threshold. A curve raising more steeply and reaching 1.0 more quickly indicates better localization performance. The evaluation is conducted on a per-scenario basis.

It can be observed that, across all scenarios, the CDF curve of the proposed multimodal fusion method converges to 1.0 the fastest, indicating superior overall accuracy and robustness. According to the CDF results, more than 95\% of the localization errors are below $0.1\,\mathrm{m}$ in simulated environments. In real-world environments, the performance slightly degrades but remains consistently high despite the challenging conditions.

\subsection{Long-term Localization Stability}
To further illustrate the robustness of the proposed MIL-LC method, a long-term localization stability test is conducted. The experiments span a period of three months, with data collected at six different time slots ranging from morning to evening. As shown in Fig.~\ref{fig:carpark_change}, the number, type, and positions of vehicles in the carpark vary over time. Typically, more vehicles are present during weekday mornings and afternoons, while fewer vehicles are observed in the evenings and on weekends. Since vehicles contain ferromagnetic materials, they can distort the local AMF. Consequently, these changes in temporal make the environment highly dynamic, leading the magnetic field distribution to vary across different data collection sessions.
\begin{table}[!t]
	\renewcommand{\arraystretch}{1.12}
	\centering
	\caption{The long-term localization accuracy under magnetic maps built on different days evaluated in Absolute Trajectory Error (ATE, m). The best result is highlighted in \textbf{BOLD}}
	\label{tab:long_term_loc_ate}
	\begin{adjustbox}{width=0.99\linewidth}
		\begin{threeparttable}
			\begin{tabular}{ccc|cccc}
				\hline\hline
				\multicolumn{3}{c|}{Configuration}                                                                                                                                                   & \multicolumn{4}{c}{Absolute Trajectory   Error (ATE, m)} \\ \hline
				\begin{tabular}[c]{@{}c@{}}Mapping\\ Date\end{tabular} & \begin{tabular}[c]{@{}c@{}}Collection \\ Time\end{tabular} & \begin{tabular}[c]{@{}c@{}}Max\\ Mag-Disturb (${\mu}T$)\end{tabular} & RoSLAC     & IDF-MFL     & PF        & Proposed          \\ \hline
				20 Dec 2025                                            & Night                                                      & 30.47                                                          & 0.145      & 0.193       & 0.159     & \textbf{0.043}    \\
				06 Feb 2026                                            & Afternoon                                                  & 80.04                                                          & 0.223      & 10.128      & 0.146     & \textbf{0.043}    \\
				07 Feb 2026                                            & Morning                                                    & 18.90                                                          & 0.120      & 16.976      & 0.137     & \textbf{0.043}    \\
				18 Mar 2026                                            & Night                                                      & 28.50                                                          & 0.145      & 0.198       & 0.147     & \textbf{0.043}    \\
				19 Mar 2026                                            & Afternoon                                                  & 53.73                                                          & 27.033     & 55.905      & 0.187     & \textbf{0.043}    \\
				20 Mar 2026                                            & Afternoon                                                  & 0.00                                                           & 0.112      & 0.117       & 0.119     & \textbf{0.043}    \\ \hline\hline
			\end{tabular}
			\begin{tablenotes}
				\footnotesize
				\item[1] The maximum disturbance is compared with map data collected on 20 Mar 2026.
			\end{tablenotes}
		\end{threeparttable}
	\end{adjustbox}
	\vspace{-1em}
\end{table}

To further visualize these variations in the magnetic field, all collected magnetic maps generated using the method in \cite{lyu2024s} are compared against a reference map collected on $20$ March $2026$. The comparison is performed by querying the predicted magnetic field $\mathbf{B}_{pred}$ from the reference map at the grid locations of other datasets, and then comparing it with the corresponding measured map values $\mathbf{B}_{map}$.
\vspace{-0.5em}
\begin{equation}
	\label{eq:diff_map}
	\small{
	B_{diff} = \|\mathbf{B}_{pred} - \mathbf{B}_{map}\|
}
\vspace{-0.5em}
\end{equation}

\begin{figure}[!t]
	\centering
	\includegraphics[width=0.999\linewidth]{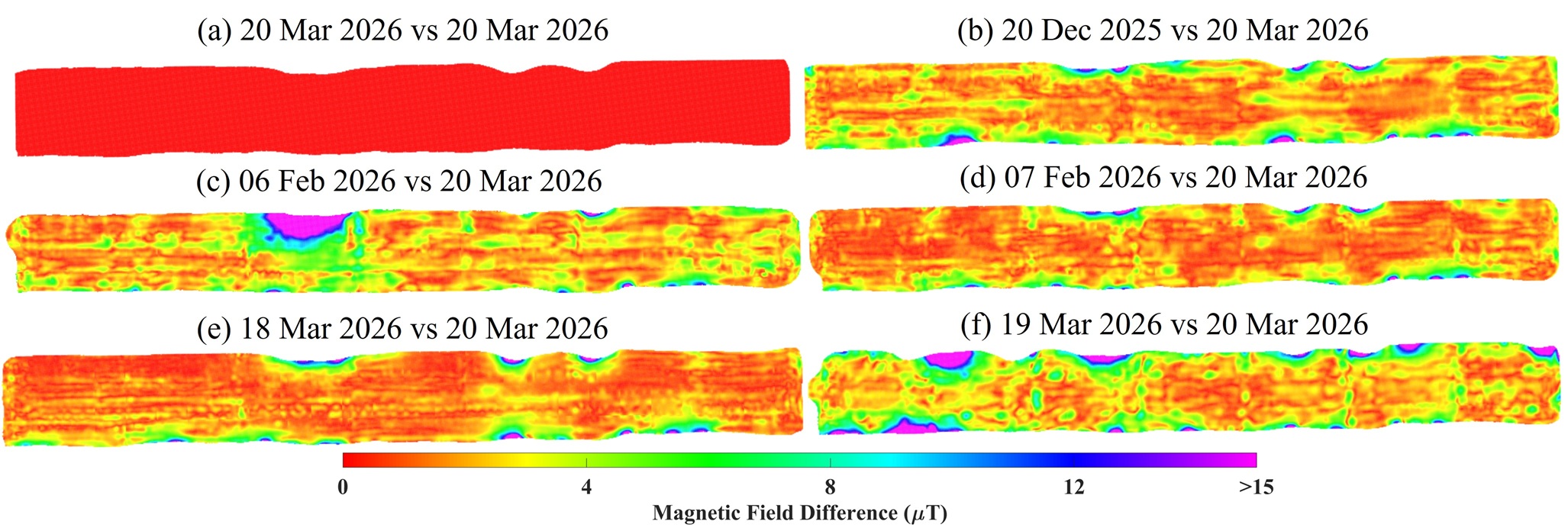}
	\caption{An illustration of the effect of vehicles on disturbances in the carpark magnetic field. The red regions indicate areas with minor disturbance, whereas the purple regions correspond to areas with severe disturbance. It can be observed that regions in closer proximity to vehicles exhibit stronger magnetic disturbances.}
	\label{fig:mag_diff}
	\vspace{-2em}
\end{figure}

The comparison results are illustrated in Fig.~\ref{fig:mag_diff}. The color encodes the difference in total magnetic intensity computed using Eq.~\ref{eq:diff_map}, where red indicates negligible difference (as expected when comparing the map of $20$ March $2026$ with itself), and blue or purple indicates significant discrepancy. From the figure, it is evident that regions surrounding parked vehicles exhibit larger differences, with the maximum variation reaching $80\mu T$ in the map collected on $06$ Feb $2026$, while central areas experience relatively smaller disturbances. Additional results are summarized in Table~\ref{tab:long_term_loc_ate}.

Based on these observations, a testing trajectory is designed, as shown in Fig.~\ref{fig:exp_traj}(f), which primarily traverses regions with significant magnetic disturbances, thereby evaluating system robustness. Specifically, magnetometer measurements collected on $20$ March $2026$ using the sensor array mounted on the Scout Mini are used, while the pre-built maps for localization are taken from different dates within the preceding three months. The four aforementioned methods are then applied for localization using these maps. The ground-truth trajectory is generated using CTE-MLO~\cite{cte_mlo} through relocalization on a Leica total-station-based map. Finally, the ATE is computed and reported in the right columns of Table~\ref{tab:long_term_loc_ate}.

From Table~\ref{tab:long_term_loc_ate}, it can be observed that the last row, corresponding to localization on the pre-built map collected on $20$ March $2026$, generally achieves the best performance across all four methods. This is expected, as both mapping and localization data are collected within a short time interval, resulting in minimal discrepancy between the magnetic map and the measurements, thereby ensuring high localization accuracy. However, when older pre-built maps are used, the accuracy of purely magnetic-based localization degrades. This degradation becomes more pronounced for maps with larger differences, such as $19$ March $2026$, where two of the compared methods fail to produce valid localization results. In contrast, the proposed multimodal fusion method demonstrates negligible degradation in localization accuracy, maintaining an ATE of approximately $0.043\,\mathrm{m}$. This highlights the advantage of the proposed approach in achieving robust and accurate localization under significant magnetic map disturbances over long-term deployment.

\begin{figure}[!t]
	\centering
	\includegraphics[width=0.999\linewidth]{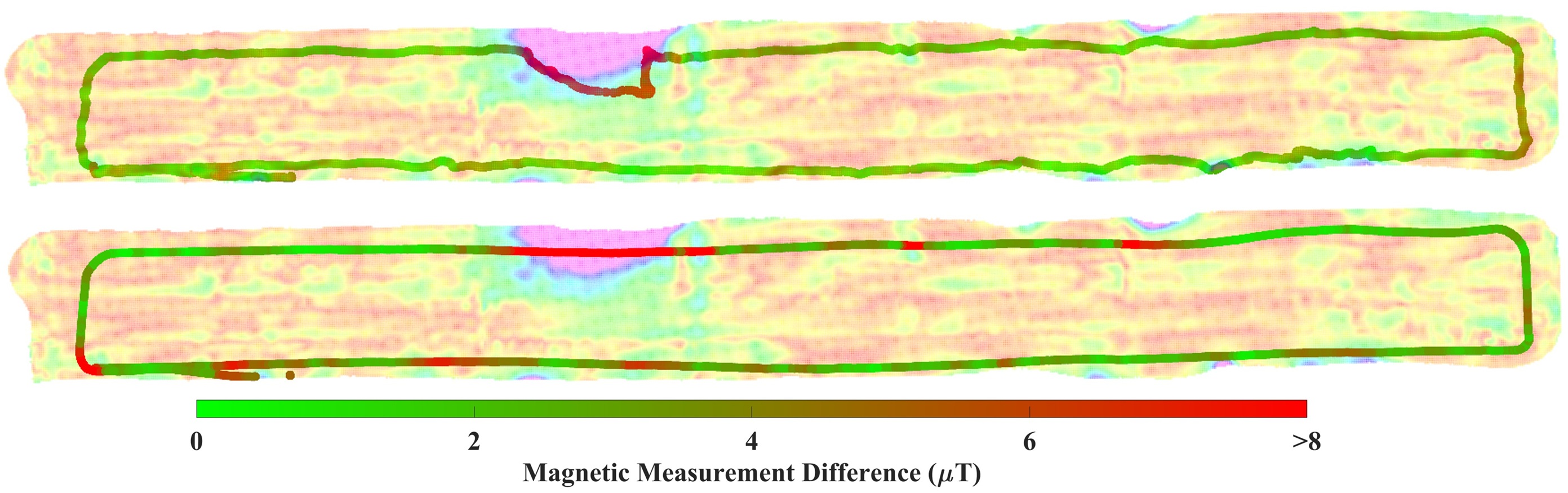}
	\caption{An illustration of the capability to highlight magnetically disturbed regions. In the top figure, pose estimation is performed using magnetic-only localization. The bottom figure employs the proposed multimodal localization framework for pose estimation.}
	\label{fig:meas_diff}
	\vspace{-1em}
\end{figure}

Given the stable accuracy over long-term localization, the proposed method demonstrates the capability to detect disturbances in the magnetic map. As shown in Fig.~\ref{fig:meas_diff}, the background field visualizes the magnetic discrepancy obtained by directly comparing two prebuilt magnetic maps (the map from 20 Mar 2026 compared with that from 06 Feb 2026). The overlaid colored trajectory represents the current absolute magnetic error, computed using Eq.~\ref{eq:diff_map}, where the online magnetometer measurements are compared with the magnetic values predicted from the prebuilt map at the corresponding positions. The red part of trajectory denotes regions with larger differences, and purely magnetic-based localization becomes unreliable; whereas green denotes regions with smaller differences, where localization remains stable. The bottom figure, which employs the fused multimodal localization trajectory, provides a more precise delineation of disturbed regions while maintaining stable pose estimation. In contrast, the top figure, based on magnetic-only localization, is less effective in highlighting these disturbed areas. This is expected, as magnetic-only localization inherently minimizes the discrepancy between measurements and the map, even when the underlying map has been perturbed.

\section{Conclusion}\label{sec:chapter6/conclusion}
This article presents a multimodal localization framework MIL-LC that fuses information from magnetometers and LiDAR as well as inertial measurements to achieve reliable localization in challenging degenerate and disturbed environments. The framework first establishes a unified approach to synchronize and integrate measurements from both sensors. To balance the high accuracy of LiDAR and the robustness of magnetic sensing, several strategies are incorporated, including degeneracy detection, outlier rejection, and adaptive weighting. Extensive experiments conducted in both simulated and real-world environments demonstrate that the proposed method achieves high reliability and accuracy across a variety of degenerate and challenging scenarios, highlighting the advantages of the multimodal fusion approach.


\bibliographystyle{ieeetr}
\bibliographystyle{./bibliography/IEEEtran}
\bibliography{mybib}

 \vspace{-3em}

\begin{IEEEbiography}[{\includegraphics[width=1in,height=1.25in,clip,keepaspectratio]{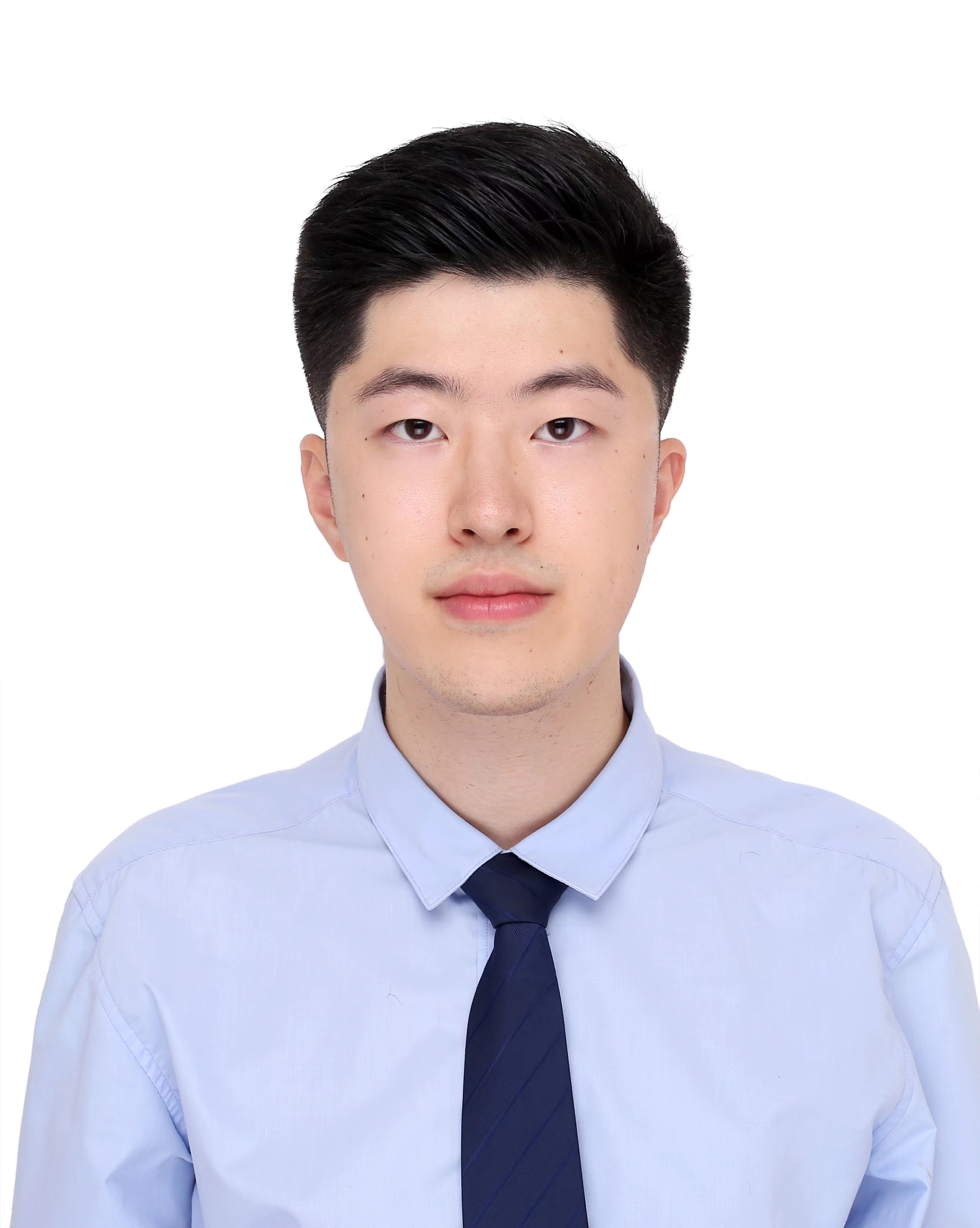}}]{Qiyang Lyu}
	received his B.Eng degree of Electronic Information Engineering from University of Electronic Science and Technology of China, China, in 2020, and the M.Sc. degrees of Computer Control and Automation from Nanyang Technological University, Singapore, in 2021. Now he is pursuing the Ph.D. degree with the School of Electrical and Electronic Engineering, Nanyang Technological University, Singapore. His research interests include sensor calibration, multi-modal mapping, and localization for autonomous robots in real complex scenarios.
\end{IEEEbiography}
\vspace{-3.2em}
\begin{IEEEbiography}[{\includegraphics[width=1in,height=1.25in,clip,keepaspectratio]{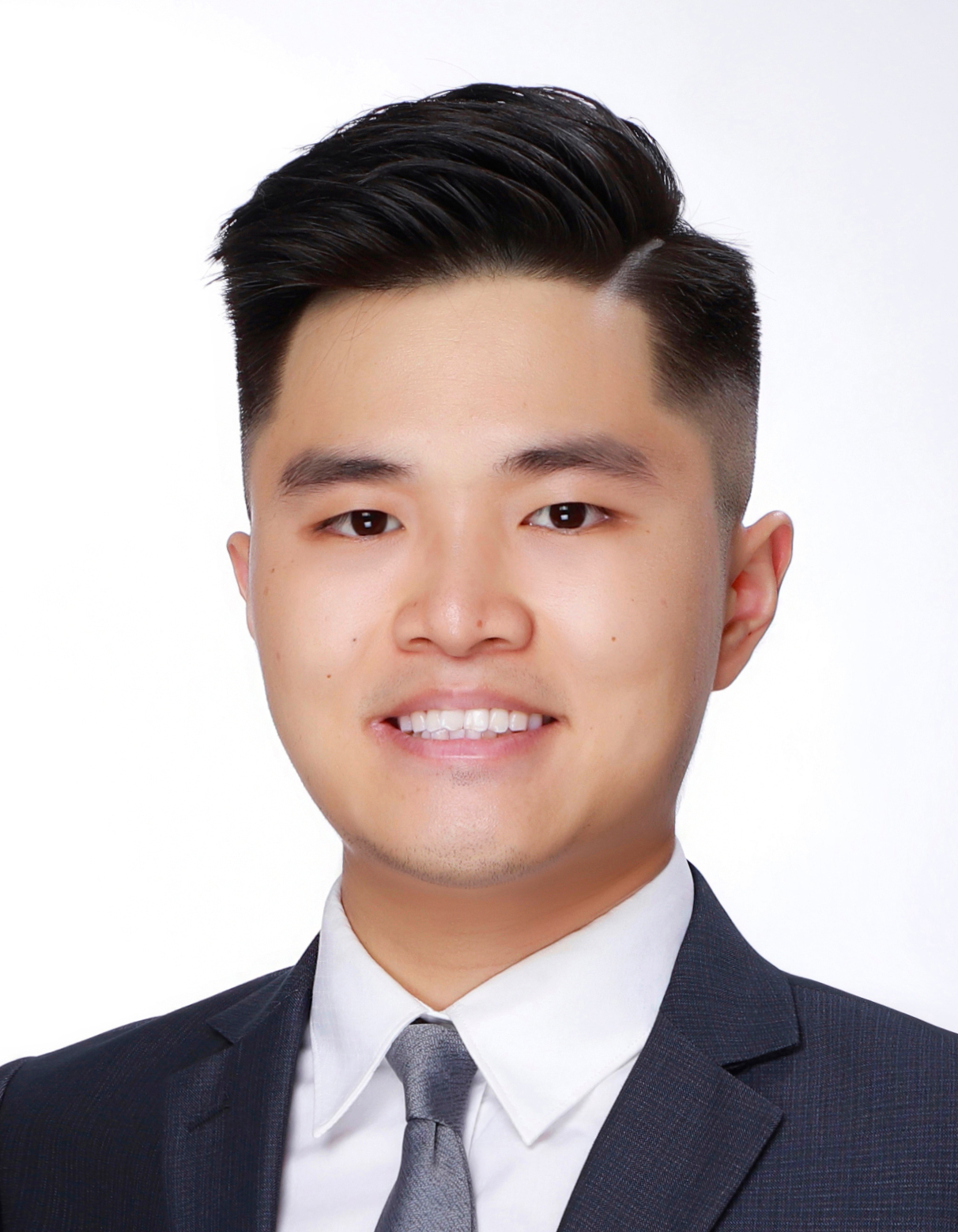}}]{Zhenyu Wu}
	received his B.Eng. degree of Electrical Engineering and Automation from Wuhan University, China, in 2016, the M.Sc. and Ph.D. degrees from Nanyang Technological University (NTU), Singapore, in 2017 and 2022, respectively. He is currently a Research Assistant Professor with the Centre for Advanced Robotics Technology Innovation (CARTIN), NTU. He served as an Associate Editor for the IEEE/RSJ IROS in 2025 and is serving as an editorial board member for Robot Learning. His research interests include intelligent perception, localization, and navigation for autonomous systems in complex environments. 
\end{IEEEbiography}
\vspace{-3.2em}
\begin{IEEEbiography}[{\includegraphics[width=1in,height=1.25in,clip,keepaspectratio]{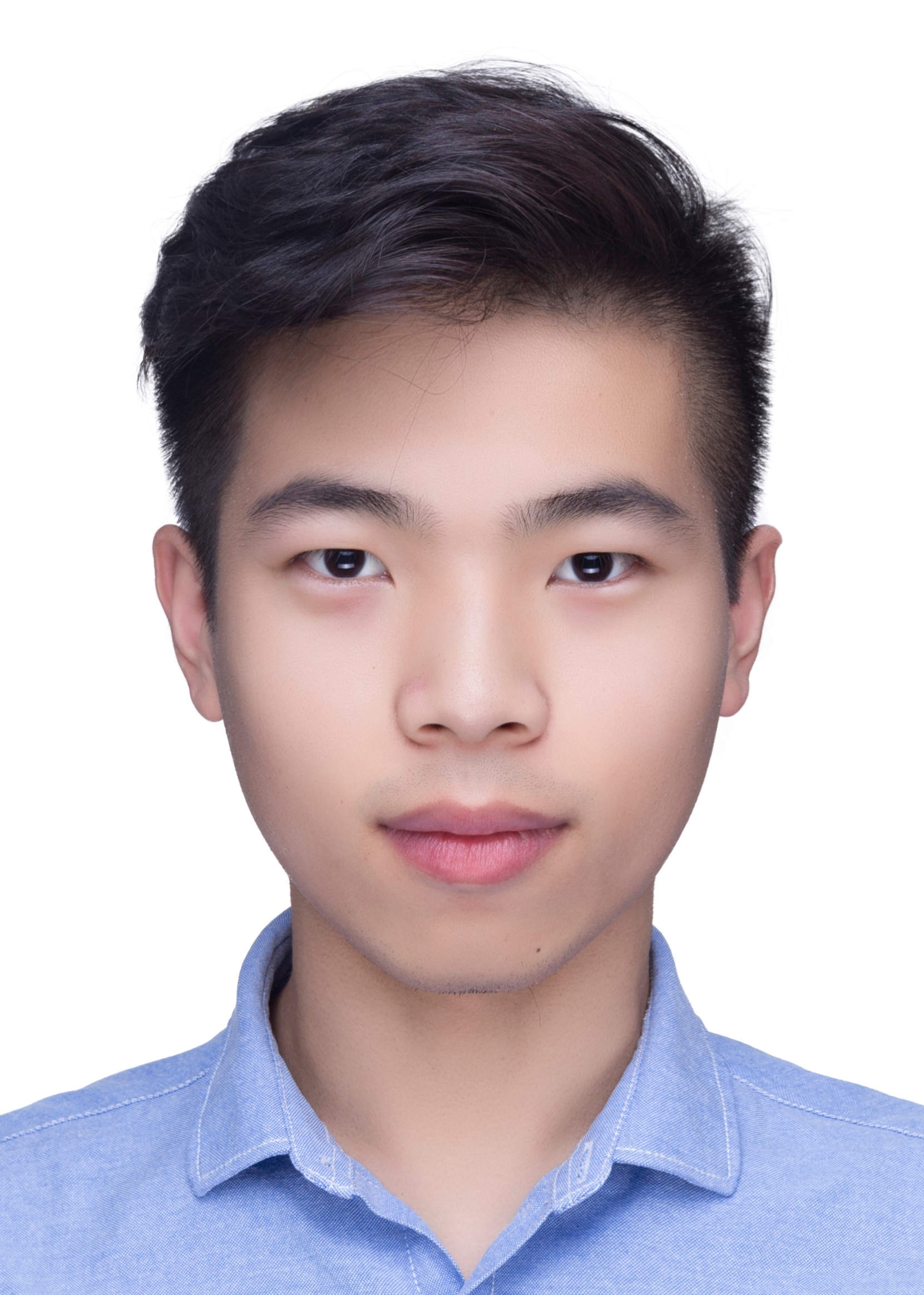}}]{Wei Wang}
	received his B.Eng degree of Automation from Zhejiang University, China, in 2020, and the M.Sc. degrees of Computer Control and Automation from Nanyang Technological University, Singapore, in 2021. Now he is pursuing the Ph.D. degree with the School of Electrical and Electronic Engineering, Nanyang Technological University, Singapore. His research interests include autonomous robots localization and navigation in complex terrains, and intelligent transportation systems.
\end{IEEEbiography}
\vspace{-3.2em}
\begin{IEEEbiography}[{\includegraphics[width=1in,height=1.25in,clip,keepaspectratio]{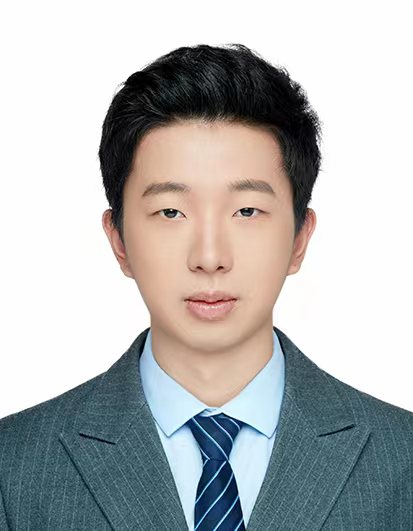}}]{Hongming Shen}
	received the B.S. degree in Flight Vehicle Design and Engineering from Central North University, Taiyuan, China, in 2015, and the M.S. degree in Aerospace Transportation and Control from the Beijing Institute of Technology, Beijing, China, in 2017. He received the Ph.D. degree in Control Theory and Control Engineering from Tianjin University, Tianjin, China, in 2023. He is currently a Postdoctoral Research Fellow with the Centre for Advanced Robotics Technology Innovation (CARTIN), NTU. His current research interests include state estimation, multisensor fusion, localization and mapping, and aerial Robotics.
\end{IEEEbiography}
\vspace{-3.2em}
\begin{IEEEbiography}[{\includegraphics[width=1in,height=1.25in,clip,keepaspectratio]{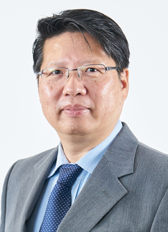}}]{Danwei Wang} (Life Fellow, IEEE) received his Ph.D. and M.S.E. degrees from the University of Michigan, Ann Arbor, USA, in 1989 and 1984, respectively. He received his B.E degree from the South China University of Technology, China, in 1982. Since 1989, he has been with the School of Electrical and Electronic Engineering, Nanyang Technological University, Singapore. Currently, he is Emeritus Professor and was the Director of the ST Engineering-NTU Robotics Corporate Lab. He is the Chair of IEEE Singapore Robotics and Automation Chapter and a senator in NTU Academics Council. He has served as general chairman, technical chairman and various positions in international robotics and control conferences, such as ICRA, IROS, and ICARCV conferences. He was a recipient of Alexander von Humboldt Fellowship, Germany, and ST Engineering Distinguished Professor Award, Singapore. He is a Fellow of the Academy of Engineering, Singapore (SAEng) and Life Fellow of the IEEE. His research interests include robotics, control theory and applications. 
\end{IEEEbiography}

\end{document}